\documentclass[10pt,journal]{IEEEtran}
\usepackage{amsmath}
\usepackage{amssymb}
\usepackage{amsthm}
\usepackage{graphicx}
\usepackage{mathtools}
\usepackage[ruled, vlined, linesnumbered]{algorithm2e}
\usepackage{color,soul}
 \usepackage{booktabs}

\usepackage{cite}
\usepackage{subfigure}
\usepackage{courier}
\usepackage{stmaryrd}
\usepackage{breqn}
\usepackage{stackengine}
\usepackage{verbatim} 
\usepackage{glossaries}
\usepackage{color,soul}
\usepackage{xcolor}
\usepackage{url}

\usepackage{xspace}

\newcommand{\nonl}{\renewcommand{\nl}{\let\nl\oldnl}}% Remove line number for one line
\makeatletter
\newcommand\approxsim{\mathchoice
  {\@approxsim {\displaystyle}      {1ex} }
  {\@approxsim {\textstyle}         {1ex} }
  {\@approxsim {\scriptstyle}       {.7ex}}
  {\@approxsim {\scriptscriptstyle} {.5ex}}}
\newcommand\@approxsim[2]{%
  \mathrel{%
    \ooalign{%
      $\m@th#1\sim$\cr
      \hidewidth$\m@th#1.$\hidewidth\cr
      \hidewidth\raise #2 \hbox{$\m@th#1.$}\hidewidth\cr
    }%
  }%
}
\usepackage{etoolbox}

\let\mybibitem\bibitem
\renewcommand{\bibitem}[1]{%
  \ifstrequal{#1}{nature}
    {\color{black}\mybibitem{#1}}
    {\color{black}\mybibitem{#1}}%
}

\makeatother

\ifCLASSINFOpdf
 
\else
  
\fi

\begin{document}

\title{A Data-Driven Framework for Improving Public EV Charging Infrastructure: Modeling and Forecasting}
\author{Nassr Al-Dahabreh, Mohammad Ali Sayed, Khaled Sarieddine, \\ Mohamed Elhattab, Maurice Khabbaz, Ribal Atallah, Chadi Assi\vspace{-0.5cm}}

\maketitle
\IEEEpeerreviewmaketitle

\begin{abstract}
    This work presents an investigation and assessment framework, which, supported by realistic data, aims at provisioning operators with in-depth insights into the consumer-perceived Quality-of-Experience (QoE) at public Electric Vehicle (EV) charging infrastructures. Motivated by the unprecedented EV market growth, it is suspected that the existing charging infrastructure will soon be no longer capable of sustaining the rapidly growing charging demands; let alone that the currently adopted ad hoc infrastructure expansion strategies seem to be far from contributing any quality service sustainability solutions that tangibly reduce (ultimately mitigate) the severity of this problem. Without suitable QoE metrics, operators, today, face remarkable difficulty in assessing the performance of EV Charging Stations (EVCSs) in this regard. This paper aims at filling this gap through the formulation of novel and original critical QoE performance metrics that provide operators with visibility into the per-EVCS operational dynamics and allow for the optimization of these stations' respective utilization. Such metrics shall then be used as inputs to a Machine Learning model finely tailored and trained using recent real-world data sets for the purpose of forecasting future long-term EVCS loads. This will, in turn, allow for making informed optimal EV charging infrastructure expansions that will be capable of reliably coping with the rising EV charging demands and maintaining acceptable QoE levels. The model's accuracy has been tested and extensive simulations are conducted to evaluate the achieved performance in terms of the above-listed metrics and show the suitability of the recommended infrastructure expansions.
\end{abstract}

\begin{IEEEkeywords}
     Charging, EV, Infrastructure, Metrics, Performance, QoE.
\end{IEEEkeywords}

\section{Introduction}\label{sec: Introduction}

\subsection{Preliminaries:} 

As the Electric Vehicle (EV) market continues to experience significant growth, there is a pressing need to expand the EV charging infrastructure accordingly. This infrastructure should ensure that the increasing demand for charging stations (CSs) is fulfilled while maintaining a satisfactory quality of experience (QoE) for users. In 2021 alone, global EV sales have doubled, bringing the total number of EVs on the road to approximately 24 million, which is three times higher than the number in 2019. The number of available public charging stations has also tripled to approximately 2 million. It is clear that the future growth of the EV market will depend on the affordability and accessibility of public or private EV charging infrastructure to the general public.

Indeed, the exponential rise in the number of EVs on the road is accelerated by efforts and investments by countries worldwide and incentives offered by governments to phase out Internal Combustion Engine (ICE) vehicles by 2050. However, there is a global disparity in EV charging network growth rates, even between different cities and regions of the same country. The International Energy Agency (IEA) \cite{IEA} uses the EV-per-Charger Ratio (EVCR) and per-EV charger power (EVCP) to assess the adequacy of charging networks. From 2015 to 2021, the EVCR for China, Korea, and the Netherlands showed a quasi-constant trend of just under 10 EVs/charger, indicating that their charging infrastructure deployment kept pace with the growing number of EVs. This was not the case for the United States, where the EVCR was recorded at a high of 18 EVs/charger in 2021. Norway's EVCR was low until 2013 but rapidly grew to 34 EVs/charger in 2021. The Alternative Fuel Infrastructure Directive (AFID) \cite{AFID} recommended EU member states achieve an EVCR of 10 EVs/charger and an EVCP of 1 kW/EV by 2020. However, the EU's 2020 average EVCR was 11, increasing to 14 in 2021, with some countries such as Italy and the Netherlands outperforming others through a wide, on-demand, slow charger deployment plan. %Despite the cumulative number of deployed slow and fast chargers, the proportion of fast chargers remained low at 3\% in the Netherlands, while other countries such as Spain and Norway had much higher levels (30\% and 35\%, respectively). The largest European EV markets, including Germany, France, and the UK, do not satisfy the EU's charger availability recommendations. 
In Canada, Quebec currently has an EVCR of 13 \cite{BernardEtAl}, of which only 10\% is covered by public chargers using both slow and fast chargers. The EVCR is expected to increase rapidly to 17 in 2025 and 29 in 2030, given Quebec's target of one million EVs on the road by 2030 and a complete ban on ICE vehicle sales in 2035. Numerous third-party initiatives, such as the Zero Emission Transportation Association (ZETA) and its members, including Uber and Tesla, support the proliferation of EVs. 

While the growth of the EV market has increased the need for a reliable public EV charging infrastructure, the current ad hoc approach to deploying additional charging stations cannot ensure proper long-term control of critical QoE metrics or instill confidence in the infrastructure. Proper tracking of events and occurrence times is necessary to provide operators with visibility into the EV charging system dynamics, allowing for accurate forecasting and optimal infrastructure expansion strategies \cite{KabirEtAl} to meet the rising demand and maintain acceptable QoE levels. Indeed, the challenges above are only a part of the many obstacles hindering the development and expansion of a reliable public EV charging infrastructure and this paper focuses on designing and developing such a framework. However, before delving into the framework, it is important to discuss first the limitations of the currently used QoE evaluation methods.

\subsection{Current QoE Metrics, Problem Statement and Motivation:} \label{sub-sec: ProbStateMotiv}
\begin{figure}
	   \centering
          %\scriptsize
	   \includegraphics[scale=0.9]{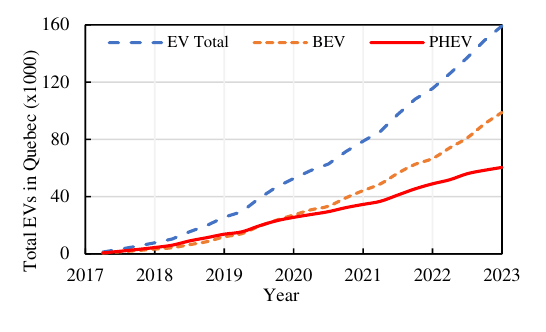}
	   \caption{EV market penetration in Quebec.}
	   \label{EV_adoption}
    \end{figure}

Ever since 2010, the worldwide achieved average EVCR has been fluctuating at slightly under $10$ EVs/charger \cite{IEA}, though, despite its globally recognized eloquence, several countries, as indicated in Section I-A have alarmingly scored two-to-three-fold this value. Keep in mind here that the main objective, now, is to show the inadequacy of the employed EV charging infrastructure expansion strategies in the majority of countries around the globe as well as the inappropriateness of the metrics that have been globally adopted to measure and indicate whether or not such expansions are being capable of coping with the rapid growth in EV market penetration rates. For this purpose, real-world data is required to ensure the correctness and credibility of the provisioned insights. Hence, following a legal agreement and officially approved research collaboration with Hydro-Quebec (HQ) \footnote{HQ owns/operates, through a subsidiary, around $80\%$ of the public EV charging infrastructure in Quebec, Canada.}, this work taps into HQ's complete database encompassing realistic daily records (overall $6$ million) of a wide range of EV charging-related measurements pertaining to comprehensive aspects of EV charging stations' dynamics (\textit{e.g.}, start time and duration of each charging session, per-session average drawn power and total drawn energy, initial installation date pertaining to each EVCS, per-EVCS nominal charging rate and geographical location)\footnote{It is worthwhile noting here that HQ's database contains no information about the per-EV arrival time and waiting time until it starts receiving service. Such are vital QoE metrics that shall be intelligently derived herein using the available records. Such metrics will allow for individual per-EVCS performance evaluation as explained hereafter.} for the various sites within Quebec (\textit{i.e.}, around $7,454$ stations distributed over $3,878$ different locations) and spanning the past five years from 2018 to 2022. With access to such a panoptic database of EV charging records, it becomes easy to evaluate and graphically visualize (as shown hereafter in Figures \ref{EV_adoption} through \label{Ratios}) the evolution of Quebec's EV charging infrastructure using the same metrics adopted by the IEA. Curves illustrated in these figures shall serve as tangible proofs of the inappropriate QoE evaluations and, hence, the ill-suited EV charging infrastructure expansions they inspire.

    \begin{figure}
	   \centering
         %\scriptsize
	   \includegraphics[scale=0.9]{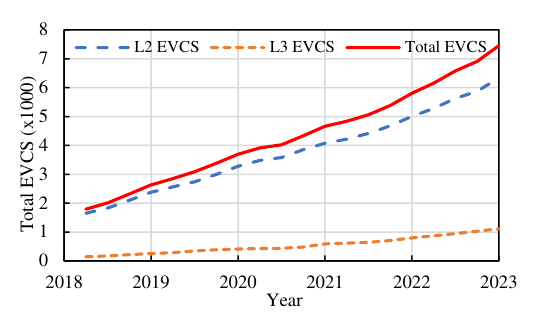}
	   \caption{Number of deployed chargers in Quebec}
	   \label{EVCS_deployment}
    \end{figure}
    
Figure \ref{EV_adoption} shows a steadily increasing total number of EVs over the entire road network of Quebec starting from year 2017 until this present time. This induces expectations of further expansion of the EV charging infrastructure at least at a rate that can ensure such an infrastructure copes with the increasing EV market penetration rate; hence, maintaining proper consumer-perceived QoE levels. In addition, Figure \ref{EV_adoption} also shows that starting from year 2019, the number of fully Battery-powered EVs (BEVs) started to overshadow that of Plug-in Hybrid EVs (PHEVs). This is a major change that promotes an increased need for more frequent per-EV charging activities as well as the need for additional energy supply to cope with this increased energy demand. This is not to mention that BEV owners would surely highly appreciate faster charging processes with less waiting; this being an expectation originating from an innate desire of QoE equity with ICE vehicles' drivers. As such, intuitively, one's thoughts get immediately directed to an increase in the deployment of advanced Level 3 (L3) fast chargers concurrently to the expansion of existing EVCSs through the addition of more of the typical Level 2 (L2) chargers. To this end, Figure \ref{EVCS_deployment} presents the EV charging infrastructure expansions observed over the past five years in Quebec; those being mainly characterized by the total numbers of L2 and L3 chargers that have been deployed and became operational and accessible by EV drivers. Here, observe that the number of L2 chargers increased by almost $300\%$ from 2018 to 2022 whereas the number of L3 fast chargers shows a relatively shy ten-fold increase. Regardless though, one cannot deny that major efforts and investments have been and continue being made to improve the availability and accessibility to a QoE-acceptable EV charging infrastructure. Unfortunately however, such efforts, are mislead to go in the wrong direction. This is especially true since they are unable to throttle down and control Quebec's continuously increasing EVCR with rising slopes as shown in Figure \ref{Ratios}. AS a matter of fact, Quebec's EVCR today scored $21$ EVs/charger. Factoring in the chargers deployed by other operators would slightly bring down the province's EVCR to $17$; this being way above the world's global average. In parallel, look at the achieved EVCP in Quebec; this being computed, following IEA's guidelines, as the ratio of the total average per-EV charging power for all available public chargers to the total number of EVs on the road. Here, note that the globally achieved EVCP amounts to $2.4$ kW/EV. Figure \ref{Ratios} shows how Quebec's EVCP has been drastically decreasing over time for it to stabilize at around $0.73$ kW/EV towards the end of 2022. At this point, also, when factoring in the contributions of the chargers pertaining to other operators in Quebec, the province's EVCP would only rise to $0.91$ kW/EV.

The above constitutes a tangible proof that the above-adopted metrics and EV charging network expansion strategies are ill-developed and that there is urgent need to establish and implement new plans for expanding this infrastructure in such a way for it to, first, fill in the gap and cope with the remarkable rise in EV adoption and, second, steadily parallel this (expected) long-term rise and meet the increasing EV charging demands both in terms of frequency and QoE-driven performance. Obviously, in light of the ealier-presented discussion, EVCR and EVCP cannot serve the general purpose of assessing the suitability of the existing infrastructure and its state-of-the-art expansion schemes as they hide from operators the factors underlying: \textit{i}) the non-homogeneous variations of EV adoption trends corresponding to different areas within the same country or province (as revealed by thorough analysis of realistic data), \textit{ii}) the contrasted location-restrictive EVCS deployment feasibility, as well as, \textit{iii}) the disparity in charging demand levels experienced in different locations. For instance, the number of EVCSs deployed in a certain touristic area might appear to be satisfactory. Keep in mind though that in such an area, often, a large proportion of the EVs requesting to charge at that area's local EVCSs may happen to be incoming from other areas. IEA's adopted metrics, namely EVCR and EVCP, do not account for these EVs when evaluated for their areas of origin. It is, therefore, of notable importance to develop adequate real-world-data-driven and location-aware EVCS QoE performance metrics that can provision insights into the suitability of existing location-dependent EV charging networks and appropriate such networks' expansion schemes.

\begin{figure}
	   \centering
          %\scriptsize
	   \includegraphics[scale=0.8]{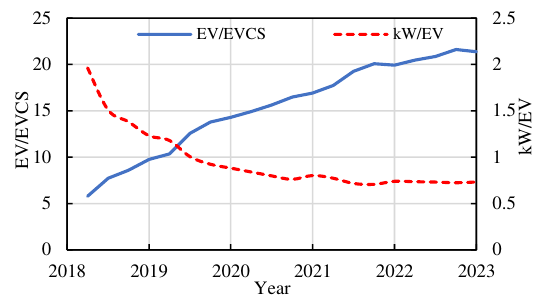}
	   \caption{Quebec's achieved EVCP v.s. EVCR.}
	   \label{Ratios}
    \end{figure}

This paper proposes a new set of critical EVCS QoE performance metrics, namely: \textit{i}) the per-EVCS achieved maximum and average waiting times, \textit{ii}) the per-charger occupancy and \textit{iii}) the per-EVCS blocking probability. The design and quantification of all of these metrics is based on HQ's comprehensive database of realistic EV charging-related records. Truly, the proposed metrics herein prevail where their existing counterparts fail at provisioning operators with visibility and insights into event dynamics of EVCSs; thus, guiding them in formulating and implementing adequate well-informed EV network sizing and charger deployment plans coupled with optimal scheduling algorithms that aiming at improving the end-consumer-perceived EV charging QoE in any particular location. To gauge their merit, a charging load forecast model adopting these new metrics is presented hereafter to predict the future charging demand per EV charging site. The next subsection highlights this paper's fundamental novel contributions and summarizes its organization. 

\subsection{Novel Contributions and Paper Organization:}

This paper aims at filling the identified literature gap consisting of the non-existence of proper QoE evaluation metrics, and the lack of accurate long-term EV charging demand forecast tools. Its contributions are briefed as follows:

$1$) To the extent of the authors' awareness, this current work is the first to present a comprehensive data-driven framework targeting the per-EVCS QoE assessment. A number of metrics are formulated for the purpose of  aiding the EVCS infrastructure operator gain a deeper understanding of their charging infrastructure utilization. These metrics shall provision EVCS infrastructure operators with in-depth visibility into the event-driven system dynamics pertaining to any particular EVCS deployed at any arbitrary location; hence, promoting operators' understanding of their charging infrastructure utilization allowing them to optimize additional chargers' deployment, and charging network sizing.

\vspace{5pt}

$2$) Comprehensive EV-charging-related historical data is exploited and fed into a newly developed Machine Learning (ML) based algorithm for long-term EV charging demand forecast. Measurement values pertaining to the number of requests in addition to external factors impacting the charging demand are fed as inputs to this algorithm allowing it to accurately predict up to one year of future per-EVCS EV charging demand.

The remaining of the paper is organized as follows. The related work is surveyed in Section \ref{sec: Related Work} and data pre-processing is done in Section \ref{sec: preproc}. Section \ref{sec: Methodology} is dedicated to the formulation of the new QoE performance metrics as well as the presentation of the ML-based forecast model. The methodology is presented in Section \ref{sec: QoEMethodology} and Section \ref{sec: Results} lays out results and discussions on the state of utilization of $14$ sites based on the developed QoE metrics herein together with the charging demand forecasts. Finally, Section \ref{sec: Conclusion} concludes the paper.

\section{Related Work}\label{sec: Related Work}

EV charging demand forecasting has been a subject that has recently received significant attention; this being attested by the increasing number of published studies presenting different approaches and shedding the light on the goodness of the resulting forecasts. Unfortunately, however, inputs to these existing forecasting models consist of simulated data generated by restrictive simulation frameworks that cannot really account for comprehensive real-world dynamics (\textit{e.g.}, housing stock, area/location characteristics, EV characteristics, actual travel distances, population growth rates, among so many others) affecting EV charging processes. In addition, a few publications do rely on some realistic data sets that, alas, happen to be quite limited in terms of the number of recorded samples corresponding to a very limited number of attributes characterizing such processes. A prime selection of these studies is surveyed hereafter. Howbeit, it is important to keep in mind that the distinguishing features that differentiate this present work from these existing publications are: \textit{i}) the exploitation of a large database of realistic records provisioning in-depth insights into actual EV charging processes taking place all over the entire Quebec province, \textit{ii}) beyond EV charging demand, this work presents a whole new set of QoE performance metrics and an accurate model to forecast future values of these metrics with such predicted results serving as guidelines for optimal EV charging infrastructure expansion and EV-to-charger assignment and scheduling.

In \cite{UnterluggauerEtAl}, the authors surveyed existing EV charging infrastructure planning methodologies that targeted the resolution of specific challenges addressed from either a transportation system perspective or that of a distribution network or, also, a combination of both. Most of these studies were theoretical in nature or based on custom-built simulation frameworks focusing on a limited subset of EV charging aspects while overlooking a large number of important factors due to the lack of visibility into them. For instance, the vast majority of these studies commonly limited their scope to the charging infrastructure itself with a focus on installation cost optimization while neglecting end-user QoE. 

The work of \cite{VashisthEtAl} presented a multi-stage EVCS deployment planning framework in an attempt to achieve an acceptable trade-off between investment costs and peak distribution grid demands. Precisely, the authors attempted to balance between EV drivers' convenience and investors' revenues without compromising capacity constraints imposed by operators over a certain pre-determined time horizon. It appears that the key driver to this framework's business succcess is the actual phase-like expected growth of the EV traffic volume, which dictates the per-stage needed additional investments targeting EV infrastructure resizing and revenues thereof originating from EV consumers' satisfaction; all this being subject to power system constraints. However, unfortunately, the proposed mathematical model in \cite{VashisthEtAl} was, contrary to the authors' claims, quite restrictive as, first, it was based on a relatively old and non-realistic EV traffic model proposed in \cite{HafezEtAl} and its restrictive assumptions leading to fictitious and non-accurate simulated distributions of the number of EVs on the road. Second, that model adopted traffic parameter values that do not conform to EV traffic in particular but rather to the vehicular mobility of all kinds of vehicles on the roads as described by the authors' traffic data source in \cite{WRTTS}; let alone, the fact that the collected data dates since 2016 (quite outdated) and that it does not incorporate any observations pertaining to the utilization and performance of deployed charging infrastructure back then. Last but not least, the model was too simplistic, stochastic in nature and aimed at separately optimizing individual non-related metrics; hence, failing to reach an even acceptable sub-optimal trade-off among these metrics that should have been jointly optimized.

The work of \cite{YiEtAl} presented one among the very few data-driven methodologies targeting the optimization of EVCS deployment within a given geographical area. Therein, the authors leveraged the PageRank algorithm (refer to \cite{PageEtAl}), Graph Theory, geographical aspects and trip data to estimate the spacial distribution of EV charging demands. More precisely, a considered study area was subdivided into cells and each cell received a PageRank score that described that cell's appeal to EV drivers. Next, a Regression Model (RM) was adopted for the purpose of mapping the PageRank scores to actual charging demands using data from existing EVCS. This RM's results were fed as input to a Capacitated Maximal Coverage Location Model (CMCLM) for the purpose of optimizing EVCS deployments with the objective of maximizing coverage. Unarguably, this presented methodology in \cite{YiEtAl} is ingenuous reflecting the authors' remarkable technical skills in combining trajectory data (provided by Inrix\footnote{https://inrix.com/} partially extracted using probing sensors (\textit{e.g.}, cell phones and automated vehicle location sensors) blending that using Google Place API with Point of Interest (PoI) data that reflected urban context and infers executed trip purposes and finally integrating socioeconomic data and land-use information (provided by Wasatch Front Regional Council\footnote{https://wfrc.org/}) used as features of people's parking behaviors that could impact the EV charging demand. However, one concern, at this point, is the fact that the majority of the used data (\textit{e.g.}, transportation-related data) is outdated (since it dates from 2016 and 2018); let alone that such data was not solely restricted to EVs but rather an entire fleet of vehicles of different types, among which are EVs. Consequently, such data may not accurately reflect current and future trends in EV adoption and EV charging demands. This is especially true given the rapid evolution of EV adoption and charging behavior patterns. Second, it is not clear how trip purposes have been linked to EV charging demand impact. It is understandable that trajectories do indicate traffic flows in and out of specific regions, of which those experiencing high traffic flows are more likely to also exhibit high EV charging demands and, hence, may constitute good locations for new charging stations. Accordingly, regardless of its purpose, an executed trip will contribute to traffic flow variations of its outbound origin and its inbound destination. In this sense, regional clusters with high connectivity and traffic flows may, to a certain extent be considered as appropriate locations for further EVCS deployments. Third, the authors presented a complex chain of interconnected machine learning models to capture the argued multifaceted nature of EV charging demand and optimal placements of new charging stations based on a variety of factors. While, to this end, the complexity of this model appears to be a point of strength, it can also set to be a limitation of this study when it comes to interpretability, transparency, credibility and utility of generated results; let alone their accuracy. The lack of knowledge and visibility into of the existing charging stations' dynamics and, hence, the per station achieved QoE performance, in a way or another, truncates future demand expectations and, hence, alters the optimality of the charging infrastructure resizing. This drawback could not, however, be worked around by the authors of \cite{YiEtAl} given not just the difficulty but, often, the rather impossibility of accessing public EV charging stations utilization information without due authorization from such stations' operators/owners. This is an issue that this present work does not suffer from given our pre-authorized access to such information.

The work of \cite{OrzechowskiEtAl} presents a data-driven management framework for EV charging stations with the objective of allowing operators/consumers to plan for peak charging times and, hence, avoid congestion. Although the authors criticize some existing short-term EV demands forecasts, their work does not serve the purpose of filling such a gap. This is especially true since their deep and supervised machine learning framework only allows for only up to one-week forecasts, which, is truly far from being long-term. Also, the utilized models therein only predict the overall energy consumptions per station. Although, the authors do consider some nice features (e.g., weather conditions, distinction between regular weekdays, weekends, and holidays, etc), such are not enough to give operators/consumers an indication of a station’s load, the waiting-in-line delay to receive service, the chargers’ utilization and the possibility of blocking (i.e., inability to provide service to an arriving EV to the station). All such gaps are accounted for in this present work whereby the presented model herein is capable of providing one full year of forecasts (extendable to two years ahead) not just in terms of energy consumption but also in terms of all of the above-listed crucial metrics as a function of the expected EV penetration rate. 

In \cite{KabirEtAl}, a Charging Station Dimensioning and Placement (CSDP) framework was presented with the objective of provisioning minimum-cost fast charging infrastructure targeting the accommodation of growing EV charging demands in a metropolitan area powered by a single power grid. Through CSDP, the authors jointly accounted for EVCS placement and capacity as well as the EV charging workload distribution among available EVCS to minimize EV waiting times and reduce range anxiety. They also factored in the power distribution network's voltage sensitivity, the possible need for voltage regulators (for maintaining voltage stability) and transformers with proper rating (for supporting peak demand). The above CSDP problem was formulated into an Integer Linear Program (ILP) characterized by its remarkable complexity that the authors worked around through the development of two heuristics. The efforts invested in developing CSDP were, indeed, seminal; especially that very little information was present at that time about EV integration, charging demands, available EVCS and their underlying functional and operational dynamics. It was, truly, an epoch of assumptions and visions that researchers attempted to concretize through the development of theoretical models and approximations that they strived to bring as close as possible to reality. Today, available real-world data continues to rule out these assumptions (\textit{e.g.}, truncated Normal distribution associated to energy demands in \cite{KhodayarEtAl}, the Normally distributed EV batter State-of-Charge (SoC) in \cite{CaoEtAl}, the exponential EV charging time in \cite{KabirEtAl, AntounEtAl}, among others). Now, although the authors of \cite{KabirEtAl} presented enough evidence of the feasibility of modelling an EVCS as a multi-server queueing system, their CSDP Workload Assignment (CSDP-WA) and sizing ILP formulations were founded on top of a highly restrictive approximation of that EVCS model using multiple single-server queues that was later shown to reflect overly pessimistic EV waiting time performance. Consequently, forecasting future demands and EVCS performances based on such allocated workload and sizing policies can surely not serve for proper charging infrastructure expansion planning.

The work of \cite{AriasEtAl} presented a big-data driven EV charging demand forecasting model accounting for vehicular traffic volume data for both vehicles and busses as well as weather conditions in addition to other variables typically considered in other existing models (\textit{e.g.}, initial battery SoC, battery type, charging power classifications, etc). Compared to older studies (\textit{e.g.}, \cite{MuEtAl, WangEtAl, QianEtAl, LojowskaEtAl}) the authors of \cite{AriasEtAl} also fed their model with the per-vehicle starting time of the charging process and initial battery SoC, which they assumed to be accurately drawn from Gaussian distributions with distinct parameters. Despite the interesting technically insightful features of the work of \cite{AriasEtAl}, it suffered from major drawbacks, first, pertaining to the non-realistic and inaccurate distributions the authors used to model the majority of their above-listed model's variables. Second, the work restricted the charging processes to take place in residential and workplaces for regular consumer EVs and in parking stations for busses. Third and most importantly, the historical traffic volume data used to train their model pertained to all types of vehicles on highways, national routes and local roads rather than just EV data. Regardless of the fact that such data dated since 2014 (\textit{i.e.}, non-representative of today's current traffic states), the authors clearly mentioned the fact that EV traffic volume at that time was much less than that pertaining to other conventional vehicles. Yet, because of their ill-paused assumption that such vehicular traffic patterns may conform to future EV-exhibited patterns due to the expected significant EV penetration growth, their reported forecasting results seriously lack accurracy. This is especially true since the EV penetration growth patterns are way different than those of conventional vehicles (as attested by currently available data); let alone, the fact that today's EV traffic patterns continue to be affected by those pertaining to typical ICE vehicles. As a matter of fact, realistically today, roads are being populated by both EV and non-EV vehicles concurrently and the presence of various publically accessible charging stations incurs significant changes in the charging demand trends. Of course such newly impactful factors did not exist back at the time when the work of \cite{AriasEtAl} was published; hence, could not be considered back then.

In \cite{MaEtAl}, the authors proposed a hybrid LSTM neural network with the objective of merging heterogeneous features pertaining to EV charging processes and, hence, accurately predict the discrete EV charging occupancy over a well defined time horizon. The reported results therein gauged the merit of the proposed algorithm and evidence its superiority over select existing benchmarks (\textit{e.g.}, hyper-parameter search \cite{BergstraEtAl}, logistic regression \cite{LinoffEtAl}, SVM \cite{SmolaEtAl}, random forest \cite{Ho} and Adaboost \cite{FreundEtAl}). The work of \cite{MaEtAl}, indeed, aims at quantifying one fundamental metric, namely, the per-charger occupancy, proposed hereafter in this present paper. As much as it is quite insightful on a technical aspect, it suffers from several drawbacks, among which, the most important is the adopted restrictive public data that describes EV charging sessions in terms of a limited number of variables allowing the designed complex forecast model to only generate relatively accurate results for only very short-term predictions ranging from $10$ minutes to only almost $4$ hours. Beyond that, the model's complexity exponentially overshoots in terms of the number of features to be considered as well as run-time only to exhibit incremental improvement over existing benchmarks. Of course, such a model would not be suitable nor insightful aiming at planning infrastructure expansion for months (let alone a minimum of a year) ahead. The forecast models presented in \cite{AminiEtAl} and those surveyed by \cite{ManujithEtAl} suffer from exactly the same issue despite their elevated accuracy of the predicted station utilization and EV charging demands for a period of time that does not bypass three days. Those models and deep-learning-based algorithms, obviously, cannot be of any utility when it comes to long-term EV infrastructure expansion planning and provide only marginal insights into the per-EVCS QoS performance.

\section{Data Pre-processing and Preparation} \label{sec: preproc}

As already mentioned in Section \ref{sub-sec: ProbStateMotiv}, the work presented in this paper is driven by a comprehensive database of $6$ million realistic charging session records pertaining to $7,454$ EVCSs deployed in $3,878$ locations. This data has been collected from the various EVCSs over the past 5 years starting from January 2018 until December 2022. Each charging session record encompasses the session's starting time, ending time, outlet ID, station ID, station postal code, and the amount of energy drawn from each outlet, among numerous other parameters that are of less interest to the work presented in this paper. Nonetheless, a major drawback of this data set is that it lacks any information about EV arrival times to the different EVCSs as well as the amount of time spent at the station waiting to start charging. Consequently, in the absence of such crucial information, the per-EVCS QoE performance evaluation becomes quite challenging. This is carefully addressed in the remaining sections hereafter. At this point though, it is quite important to mention that after closely observing the available records, some of them can be identified as invalid. For instance, all records pertaining to sessions that: \textit{i}) lasted less than $3$ minutes or more than $2$ hours, \textit{ii}) have no recorded payments associated with them, and, \textit{iii}) show that no energy has been drawn from the corresponding outlet, are assumed to be invalid records, and, hence, have been removed. Following this cleaning phase, the remaining records have been, first, arranged in ascending order of their corresponding postal codes, and then, for each station, records are ordered according to the starting times of each session. To this end, a careful examination of the postal codes and site IDs pertaining to each station reveals the operator's attempt to cluster regions based on the postal code. Consequently, all stations having the same postal code are considered as belonging to the same site and, hence, associated with the same site ID. %This is quite interesting as it allows one to pinpoint any charging infrastructure performance anomalies both at a regional level and, to a finer granularity, at each and every EVCS belonging to any particular site. 

%% Chadi removed this section...
%Now, the above data manipulations shed the light on an exceptional epoch spanning a total of $18$ months starting from January 2020 all the way to June 2021 (hereafter referred to as the gap period) during which the public charging infrastructure, at the provincial level, experienced a severe drop in EV charging demands. One cannot consider such a drop as an anomaly as this is expected and appears as one of the dreadful consequences of COVID-19. %and the forced long lockdown periods, curfews, border closures and shifting all kinds of academic and work/industrial activities to online platforms; all of these, obviously, contributing to a dominant transport sector stagnation; hence, inevitably hindering the proliferation of EVs. 
%Figure \ref{fig: demandCOV19} illustrates the variations of EV charging demands from January 2018 to December 2022 and, clearly, reflects the severe recession experienced during the above-mentioned gap period. 

\section{EVCS QoE Performance Metrics} \label{sec: Methodology}

This section presents two sets of novel metrics for assessing the quality of service (QoS) provided by public electric vehicle charging stations (EVCS) in Quebec, Canada. The first set of metrics includes the number of charging requests, per-site utilization, per-site occupancy, idleness, and blocking probabilities, which are based on a comprehensive database of realistic measurements taken at different times and pertaining to various EV charging process variables. These metrics are subsequently used as inputs to a forecast model developed for predicting future EV charging loads. They are as follows:

\noindent $\bullet$ \textbf{Number of Charging Requests}, $N_{R}$, represents the number of unique charging requests experienced by an EVCS during a given time period. The provided data reveals that some EVs stop but then resume charging within a few seconds. In what follows, such short charging interruptions are neglected (\textit{i.e.}, the continuation of charging is not considered as a new request placed by the same EV but rather the same request identified using the account credentials of the user who initiated the charging session). Also, all charging sessions with only a few seconds durations are discarded.

\noindent $\bullet$ \textbf{Charging Site Utilization}, $U$, determines if a site is being underutilized or over-utilized. Through the observation of the variations of a site's daily instantaneous utilization, an operator can determine the exact EV charging demand at that site at any particular point in time. Also, $U$'s long-term variations reveal the direct impact of EV consumers' socio-economical habits on EV charging requirements. Consequently, $U$ provides operators with insights into the load experienced per site, which cannot be provided by $N_{R}$ alone as this latter contains no information about the different charging sessions' durations. By discretizing time into mini slots of duration $1$ minute each: 

\begin{equation}
    \label{eq:U}
    U =\frac{1}{T}\displaystyle\sum_{i} \frac{k_i}{n}
 \end{equation}

\noindent where $T$ is the total number of time slots within an observation period, $i \in \mathbb{N}$ used for indexing each time slot within $T$, $k_i$ is the number of occupied EVCSs during time slot $i$ and $n$ is the number of per-site EVCSs.

\noindent $\bullet$ \textbf{Site Occupancy}, $\Omega$, represents the probability that at least one of a given site's EVCSs is occupied (\textit{i.e.}, the probability of that charging site being utilized):
   
\begin{equation}
    \label{eq:omega}
    \Omega= \frac{1}{T} \sum_{i} \mathcal{Z}_{i}\\
       \textrm{          where $\mathcal{Z}_{i}$ =} 
    \begin{cases} 
        0\text{ if }k_{i} = 0\\
        1\text{ if }k_{i} \geq 1
    \end{cases}
\end{equation}

\noindent $\bullet$ \textbf{Site Idleness Probability}, $P_{I}$, represents the fraction of the total observation duration $T$ during which all EVCSs pertaining to a given site are idle. This metric quantifies the possibility of an EV arriving at a given site and finding all EVCSs available. As the complement of $\Omega$:

\begin{equation}
    \label{eq:P_I}
    P_I = \Pr\left[k_i = 0\right] = 1 - \Omega
\end{equation}

\noindent $\bullet$ \textbf{Site Blocking Probability}, $P_{B}$, represents the proportion of $T$ during which all of a given site's EVCSs are found to be completely busy (\textit{i.e.}, the probability of an EV arriving at a given site and finding all EVCSs occupied and, hence, suffering from immediate denial of service): 

\begin{equation}
    \label{eq:PB}
     P_{B}= \frac{1}{T} \sum_{i} \mathcal{X}_{i}\\
    \textrm{          where $\mathcal{X}_{i}$ =} 
    \begin{cases} 
        1\text{ if }k_{i} = n\\
        0\text{ if }k_{i} \leq n
    \end{cases}
\end{equation}

\noindent Beyond a certain QoE threshold, $P_B$ indicates the failure of the existing charging infrastructure to sustain the growing EV charging demands experienced by the given site.

Next, the second set of metrics includes the longest per-site busy period and average EV waiting time, which are evaluated and reported by a custom-built simulator. These metrics aim to quantify the quality of perceived service by EV consumers at different public EVCSs and can be used by EV charging infrastructure operators worldwide to assess the performance of any charging site and plan EV charging network expansions as needed. They are:

\noindent $\bullet$ \textbf{Number of Delayed EVs}, $N_{D}$, represents the number of EVs that have experienced a certain delay at a given site waiting to receive service (\textit{i.e.}, start charging). Any EV that starts charging within at most $t$ minutes (e.g., $t\leq 5$minutes) from the end of a preceding charging session at a blocked site is considered to be a delayed EV. The value of $t$ is not fixed and is empirically estimated and recorded by the operator on a per-site basis given the fact that it is affected by numerous uncontrollable factors that are external to the charging process (\textit{e.g.}, parking lots’ availability close to rest areas and shopping malls and their proximity to EVCSs, in-proximity services, etc). In the future, several solutions may be adopted to increase the accuracy of $t$’s values (\textit{e.g.}, the installation of cameras to provide video footage showing the advancement of a waiting EV into service position and, hence, the start of a new session).

\noindent $\bullet$ \textbf{Average Waiting Time}, $\overline{W}$, is the average queueing delay experienced by EVs waiting at a given site to start charging. $\overline{W}$ is generated using a custom-built PHYTON-based discrete-event simulator that models each site as an \textit{M}/\textit{G}/$k$ queue.

\begin{figure}[t]
	\centering
    \scriptsize
	\includegraphics[scale = 0.4]{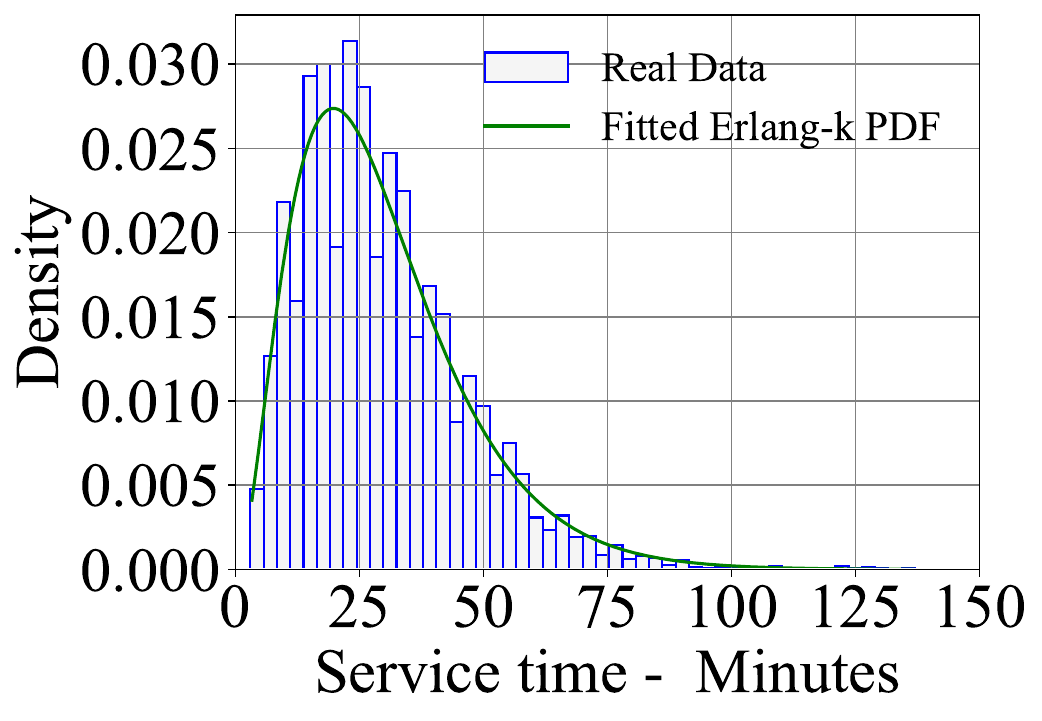}
	\caption{Sample service time data distribution.}
	\label{fig: Erlangk}
\end{figure}
    
\section{QoE Evaluation Methodology}\label{sec: QoEMethodology}

The adopted per-site QoE-oriented performance analysis methodology is presented. First, although the above-presented metrics may provide per-EVCS performance insights for the chosen site, it is observed that performance trends notably vary for different observation period categories (\textit{e.g.}, regular weekdays, weekends, holidays, etc). As such, to capture such variations, QoE performance evaluations shall be conducted on daily basis spanning any required duration (\textit{i.e.}, week, month, year). Precisely, the examined days throughout the analysis period shall be grouped based on their different categories. Then, for each one of these days, each of the above-listed metrics shall be evaluated. This will indicate the frequency whereby each of these metrics' respective values signaled a red flag for each one of these days' categories; this being of utmost importance to operators when developing EV charging infrastructure expansion plans. For instance, in touristic areas, the achieved QoE might be misleadingly perceived as good whenever values of the $U$, $\Omega$, $P_{B}$, $N_{D}$, and $\overline{W}$, do not bypass a given threshold when evaluated for one whole month or year in one shot. The operator, here, may be deluded to trust a site's operational normality whereas, truly, these metrics' daily values may reveal an overall site under-utilization during normal work days and an over-utilization (\textit{i.e.} low QoE) during weekends and holidays.

\subsection{Simulation Framework:}

A custom-built discrete-event PYTHON-based simulator is developed herein for the purpose of modelling any EVCS pertaining to any charging site as an \textit{M}/\textit{G}/$k$ queueing system. This simulator's input parameters' values are set according to simulated EVCS's information extracted from the given dataset. Precisely, EV arrivals are assumed to follow a Poisson process with a parameter $\lambda_E = N_E \cdot T^{-1}$ EVs/s with $N_E$ being the number of observed EVs arriving to the simulated EVCS within the observation period $T$. $N_E$ is dictated by HQ's recorded data. The per-EV service time is drawn from an empirically evaluated Erlang-$k$ distribution that fits HQ's service time records for the simulated sites. Due to space limitation, only one such distribution example is illustrated herein in Figure \ref{fig: Erlangk} where the Root Mean Square Error (RMSE) between the realistic distribution and its theoretically fitted counterpart is of the order of $10^{-8}$ indicating a highly accurate fit. Finally, the simulated EVCS's number of outlets is $k$ and is also provided by the operator. The fundamental objective of this developed simulator is to generate $N_D$ and $\overline{W}$. Furthermore, future EV arrival forecasts are going to be generated hereafter. Such arrival forecasts may then serve as inputs to the above-mentioned simulator, which, then, shall return forecasts for future $N_D$ and $\overline{W}$ values pertaining to a simulated EVCS.

\begin{figure}
	\centering
    \scriptsize
	\includegraphics[width=1\linewidth]{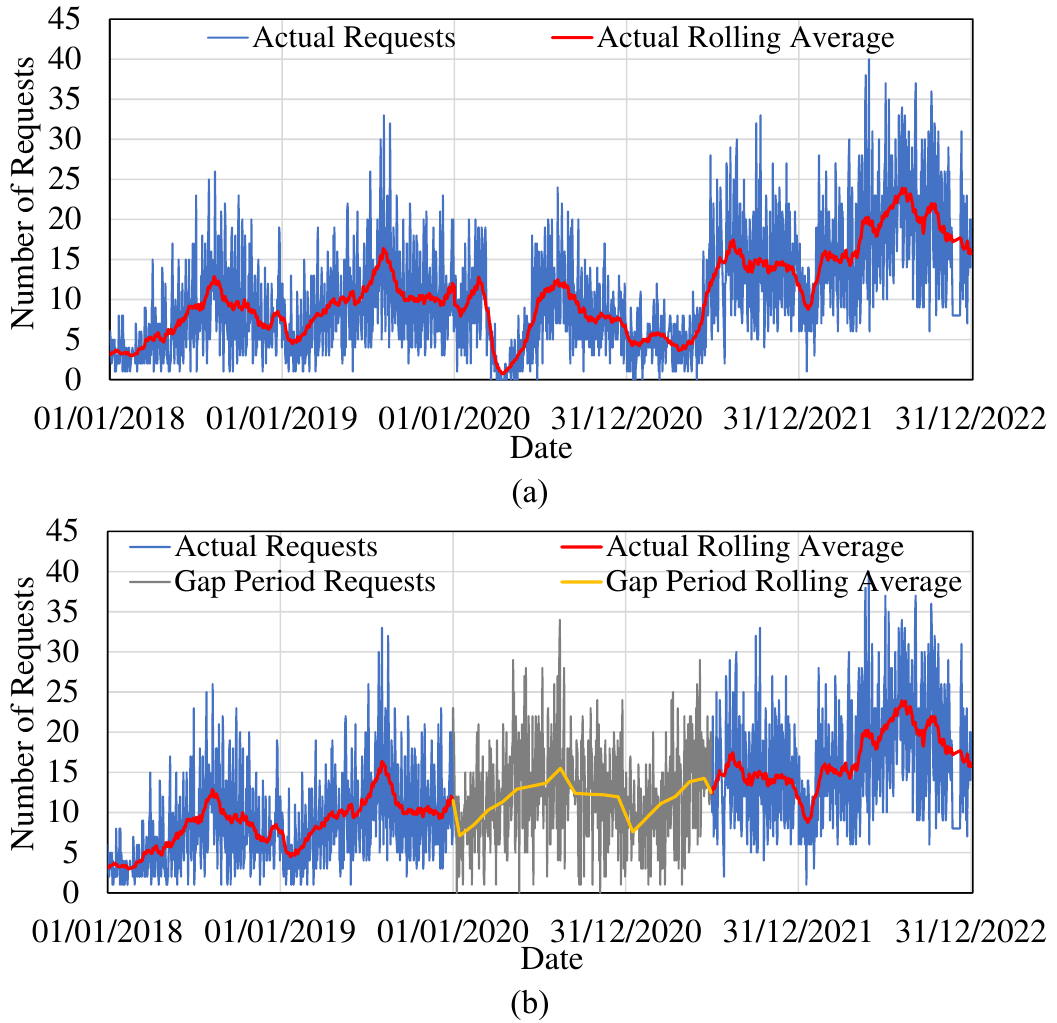}
	\caption{Five year charging requests (a) including the COVID-19 data and (b) requests after filling the gap caused by COVID-19}
	\label{fig: Gap}
\end{figure}

\subsection{Filling the Gap in the Dataset Created by COVID-19:}
 
In an attempt to combat the proliferation of COVID-19, governments around the Globe and, particularly in Quebec, imposed numerous and lengthy curfews, lockdowns and border closures. In addition, remote work and schooling policies were implemented all over the province. All these have remarkably affected the province-wide transportation sector. Precisely, during $18$ months (from January $2020$ to June $2021$) the number of vehicles on the road decreased by almost $60\%$ including the number of EVs. The decrease in this latter reflected itself in an abnormal decrease in EV charging demands as illustrated in Figure \ref{fig: Gap}(a); hence, the reason behind referring to this time period hereafter as the "\textit{gap period}". For all purposes of proper QoE performance forecasting, proper charging demand trend corrections (data augmentation) need to be applied to this gap period in order for it to appear as a smooth quasi-normal continuation to its predecessor and successor periods; that is, as if the pandemic never occurred. This is achieved as follows:

\vspace{5pt}

\noindent $1$) Among the $5$ years data records pertaining to a given site, disregard all those corresponding to the gap period.

\vspace{5pt}
    
\noindent $2$) Compute an $M$ days weighted moving average whereby, the number of requests for each target day of the gap period becomes equal to the average number of requests observed during the $\frac{M-1}{2}$ past and future days. As such, the target day is that $M^{\text{th}}$ day that appears at the center of the $M$ days period. This mathematically translates to:

    \begin{equation}\label{eq:MAv}
        \overline{N_{R, d}} =\frac{1}{M-1}\displaystyle\sum_{j=d-\frac{M-1}{2}}^{d+\frac{M-1}{2}} N_{R,j}
    \end{equation}

\noindent where $d$ is the index of the targeted gap day and $N_{R,j}$ is the actual number of requests.

\vspace{5pt}

\noindent $3$) Compute the relative daily charging request difference $\Delta N_R = \left(N_{R,d} - \overline{N_{R,d}}\right)\overline{N_{R,d}}^{-1}$. Use all relative differences over the entire $42$ months before and after the gap to generate $\Delta N_R$'s empirical probability distribution, $\mathcal{D}$.

\vspace{5pt}

\noindent $4$) Finally, to each computed $\overline{N_{R,d}}$ over the $18$-month gap period, add a random $\Delta N_R$ value drawn from $\mathcal{D}$.

Figure \ref{fig: Gap}(a) ($M = 30$) concurrently plots $N_{R,d}$ and $\overline{N_{R,d}}$ pertaining to one selected site over the entire $5$-year period. The trend anomaly is clear during the gap period compared to its normal counterparts appearing before and after the gap. Most importantly, Figure \ref{fig: Gap}(a) tangibly proves $\overline{N_{R,d}}$'s potency in capturing the seasonal/annual trends of $N_{R,d}$ while smoothing out its randomness. Figure \ref{fig: Gap}(b) shows the reconstructed number of charging requests values corresponding to the gap period following the above-elaborated procedure. These results assert the suitability of the utilized approach since the reconstructed trend appears clearly as a smooth quasi-normal bridge connecting its predecessor to its successor curves.  

\subsection{Long-Term Forecast Model:}

As anticipated, the EV charging demand is expected to grow at an increasing pace in the coming years. This stresses the importance of creating a long-term forecast model that can accurately predict the EV charging load in terms of served request counts over an entire year while taking the load’s seasonality into consideration. 
 
\noindent $\bullet$ \textbf{Data Pre-Processing and Feature Engineering:} The first pre-processing step consists of chronologically arranging the charging records to create a continuously evolving time series and then extract from them the required features, the fundamental one of which are the timestamps listed in Table \ref{table:timestamp}. Here, timestamp differentiation and proper interpretation are important especially because of the differential characteristics of regular weekdays' charging requests and their weekends/holidays counterparts. Table \ref{table:features} lists additionally extracted categorical features constituting external factors that impact the charging demand. All of these features are then encoded using the mean encoding method described in \cite{Chatfield} and \cite{encod} to reveal a logical correlation between them and their corresponding label (\textit{ i.e.}, $N_{R,d}$). Consequently, during the training phase, the model determines the relationship between the predicted value and the mean features' encoding instead of the actual $N_{R,d}$.

\begin{table}
    \centering
    \caption{Time stamp features}
    \begin{tabular}{|l|c|}
        \hline
        \hline
            \textbf{Feature Name} & \textbf{Range of Values}\\
        \hline
        \hline
            Weekday (Monday$\rightarrow$Sunday) & $[0;6]$\\
        \hline
            Day of the month & $[1;31]$\\
        \hline
            Day of the year & $[1;365]$\\
        \hline
            Month of the year & $[1;12]$\\
        \hline
            Week of the year & $[1;53]$\\
        \hline
            Week of the month & $[1;4]$\\
        \hline
            Quarter of the year & $[1;4]$\\
        \hline
            Recorded year & $[2018;2022]$\\
        \hline
            Working Day & [$0$: Weekend/Holiday; $1$: Business day]\\
        \hline
        \hline
    \end{tabular}
    \label{table:timestamp}
\end{table}

\begin{table}
    \centering
    \caption{Charging service features}
    \begin{tabular}{|l|c|}
        \hline
        \hline
            \textbf{Feature Name} & \textbf{Range of Values}\\
        \hline
        \hline
            Charging requests weighted rolling average & Float\\
        \hline
            Province-wide number of available public EVCSs & Integer\\
        \hline
            Province-wide number of registered EVs & Integer\\
        \hline
            Region-specific number of available public EVCSs & Integer\\
        \hline
            Region-specific number of registered EVs & Integer\\
       \hline
       \hline
    \end{tabular}
    \label{table:features}
\end{table}

\noindent $\bullet$ \textbf{Forecast Model:} A Seasonal Auto-Regressive Integrated Moving Average with eXogenous factors (SARIMAX) statistical learning model (\textit{e.g.}, \cite{sarimax}) is presented hereafter. This model takes as input a time series rendering it capable of accurately predicting future $N_{R,d}$ values while concurrently capturing its real-world historical data inputs' seasonality and patterns. In the sequel, it is proven that SARIMAX is capable of: \textit{i}) accurately capturing and representing this seasonality for $1$-year forecasts, and, \textit{ii}) integrating multiple external variables and deduce their impact on the charging demand's behavior.

\begin{table}[t]
    \centering
    \caption{\textsc{Charging Site Description and Number of EVCSs per Site}}
    \begin{tabular}{|c | l | c|}
    \hline
    \hline
        \multicolumn{1}{|c}{\textbf{Site \#}} & \multicolumn{1}{|c}{\textbf{Location}} & \multicolumn{1}{c|}{\textbf{Number of EVCSs/Site}}\\
        \hline
        \hline
        \textbf{1} & City 1 Downtown & 2\\
        \textbf{2} & City 1 Residential Area & 2\\
        \textbf{3} & City 2 Residential Area & 2\\
        \textbf{4} & City 2 Mall Parking Lot & 2\\
        \textbf{5} & City 3 Downtown & 1\\
        \textbf{6} & Suburb 1 & 1\\
        \textbf{7} & Suburb 2 & 2\\
        \textbf{8} & Rural Area 1 & 1\\
        \textbf{9} & Rural Area 2 & 2\\
        \textbf{10} & Touristic Area 1 & 2\\
        \textbf{11} & Touristic Area 2 & 1\\
        \textbf{12} & Highway 1 & 2\\
        \textbf{13} & Highway 2 & 2\\
        \textbf{14}& Highway 3 & 4\\
        \hline
        \hline
    \end{tabular}
    \label{table: sites}
\end{table}

As a matter of fact, the accessible data sets related to the addressed problem in this paper enclose data sample points that constitute a time series. This is especially true since these samples represent the magnitude of changes in the EV charging load as a function of time. The objective here is to forecast future metric values (and their variations as a function of time as well) according to what can be learned from the past history embedded in the above-said time series. As such, given that such data exhibits seasonality patterns and given the availability of a strong exogenous factor in the time series, SARIMAX would be the best forecasting model fit in this case.

SARIMAX is defined by three main parameters, \texttt{p}, \texttt{d}, and \texttt{q}, respectively denoting the number of auto-regressive terms, the order of differentiation, and the order of moving averages. The determination of seasonal variations is required to establish a Seasonal Auto-Regressive Function (SARF) in addition to a Non-seasonal Auto-Regressive Function (NARF) \cite{AR}. Additionally, the set of exogenous variables is fed into the model in an array-like parameter \texttt{exog}. Proper tuning SARIMAX's parameters allows for the generation of accurate forecasts based on the detected patterns and the exogenous features. The recorded daily per-site number of charging sessions and their corresponding exogenous variables are aggregated weekly to overcome $N_{R,d}$'s randomness. This kind of aggregation is acceptable and rather preferred for objectives similar to those adopted in this paper (\textit{i.e.}, forecasting a charging site's QoE future evolution).

Initially, the Box-Jenkins approach is used to estimate the most appropriate range for the \texttt{d} parameter. Then the Partial Auto-correlation function was used to determine an appropriate range for \texttt{p} and the Auto-correlation function was used to determine \texttt{q}. Augmented-Dickey-Fuller tests are executed to verify the existence of non-stationary conditions. A close examination of these tests' results reveals that the best model is one to which correspond fractional parameters' values. In fact, Fractional ARIMA is used when the data patterns exhibit long-range dependencies that cannot be captured using integer parameters. The interpretation of the fractional parameters becomes different than the interpretation of the integer parameters. Fractional differencing with \texttt{d} $< 1$, is used to smoothen out very long-term data dependencies while simultaneously preserving the short-term dependencies. A value of $0.5 <$ \texttt{d} $< 1$ would indicate the presence of a strong historical dependence in the data. Additionally, using a fractional \texttt{d} eliminates the issue of over-differencing which might add white noise when large values greater than $1$ are used. Additionally, it is favorable to choose low values of the \texttt{p} and \texttt{q} when using fractional values of \texttt{d} to learn the long-term dependencies of the data while simultaneously preserving the short-term fluctuations and dependencies. As a result, low values of the fractional auto-regressive term \texttt{p} are used to capture the short-term dependencies of the data. Finally, low values for the moving average order \texttt{q} are used to capture the short-term fluctuations and noise in the data. As a result, grid search is performed here to tune the parameters of the SARIMAX model adopted herein with low range of \texttt{p} and \texttt{q} and a fractional \texttt{d}. Additionally, the choice of exogenous variable is extremely important to improve the model's accuracy significantly. This is the number of EVs in the region (scaled by the average charging demand of a typical EV) was selected as an exogenous variable.

The model is trained and optimized against the data extending from $2018$ to $2021$ based on the above-described approach. Then, it is tested against non-training data (\textit{i.e.}, pertaining to year $2022$), to validate its ability to forecast an entire year accurately. Several accuracy metrics are used to evaluate the model's accuracy such as RSME, MSE, MAE, and MAPE.

The SARIMAX model proposed herein forecasts the weekly average number of requests with $99\%$ confidence. This allows, at this point, to utilize this model to forecast the charging requests for the selected sites throughout the entire next year $2023$. Note that the model requires re-training using the data of each individual site for accurate forecasts to be generated. Additionally, a second year forecast is generated to guide the operator with some longer-term forecasts. However, it is important to highlight that the farther future forecasts are generated with a reduced results' confidence. Yet, every new year of data that is added to the data set enables longer and more accurate the forecasts.

Now, indeed, there are so many different prediction and forecasting models. It is quite interesting to compare the performance of the adopted SAIMAX model with fractional parameters to other models such as LSTM, ETS, ARIMA, and SARIMA. This is done in Section VI below.

\section{Analysis and Forecast Results}\label{sec: Results}

This section presents an evaluation of the performance of $14$ selected charging sites in terms of the presented metrics in Section \ref{sec: Methodology}. Moreover, $4$ of these sites are selected to showcase the prediction accuracy of the developed SARIMAX. The reported results also illustrate the average waiting time evolution between $2021$ and $2022$ as well as the predicted waiting time for $2023$ highlighting the impact of the increasing EV charging demand on the user-perceived QoE.

\begin{table}[t]
    \centering
    \caption{\textsc{Charging Site Occupancy in $2021$ and $2022$}}
    \begin{tabular}{|l | c c | c c|}
        \hline
        \hline
        \multicolumn{1}{|c}{ } & \multicolumn{2}{c|}{\# Days 10\%$\leq \Omega \leq$30\%} & \multicolumn{2}{c|}{\# Days $\Omega \geq$ 30\%}\\\hline
        \multicolumn{1}{|c}{\textbf{Site Location}} & \multicolumn{1}{c}{\textbf{2021}} & \multicolumn{1}{c|}{\textbf{2022}} & \multicolumn{1}{c}{\textbf{2021}} & \multicolumn{1}{c|}{\textbf{2022}}\\
        \hline
        \hline
        City 1 Downtown & 239 & 131 & 107 & 232\\
        City 1 Residential Area & 182 & 192 & 8 & 13\\
        City 2 Residential Area & 247 & 233 & 14 & 20\\
        City 2 Mall Parking Lot & 178 & 245 & 20 & 67\\
        City 3 Downtown & 139 & 206 & 6 & 12\\
        Suburb 1 & 127 & 190 & 3 & 10\\
        Suburb 2 & 160 & 229 & 84 & 34\\
        Rural Area 1 & 170 & 198 & 52 & 113\\
        Rural Area 2 & 145 & 159 & 70 & 161\\
        Touristic Area 1 & 154 & 85 & 110 & 263\\
        Touristic Area 2 & 187 & 230 & 21 & 41\\
        Highway 1 & 91 & 147 & 5 & 6\\
        Highway 2 & 31 & 107 & 0 & 1\\
        Highway 3 & 220 & 111 & 88 & 252\\
        \hline
        \hline
    \end{tabular}
    \label{table: occupancy}
\end{table}

\begin{table}[t]
    \centering
    \caption{\textsc{Charging Site Utilization in $2021$ and $2022$}}
    \begin{tabular}{|l | c c | c c|}
    \hline
    \hline
        \multicolumn{1}{|c}{ } & \multicolumn{2}{c|}{\# Days 10\%$\leq U \leq$30\%} & \multicolumn{2}{c|}{\# Days $U$ $\geq$ 30\%}\\\hline
        \multicolumn{1}{|c}{\textbf{Site Location}} & \multicolumn{1}{c}{\textbf{2021}} & \multicolumn{1}{c|}{\textbf{2022}} & \multicolumn{1}{c}{\textbf{2021}} & \multicolumn{1}{c|}{\textbf{2022}}\\
        \hline
        \hline
        City 1 Downtown & 267 & 303 & 12 & 38\\
        City 1 Residential Area & 69 & 70 & 0 & 0\\
        City 2 Residential Area & 108 & 116 & 0 & 0\\
        City 2 Mall Parking Lot & 91 & 186 & 0 & 0\\
        City 3 Downtown & 139 & 206 & 6 & 12\\
        Suburb 1 & 127 & 190 & 3 & 10\\
        Suburb 2 & 149 & 127 & 17 & 0\\
        Rural Area 1 & 170 & 198 & 52 & 113\\
        Rural Area 2 & 135 & 187 & 17 & 63\\
        Touristic Area 1 & 165 & 210 & 27 & 123\\
        Touristic Area 2 & 187 & 230 & 21 & 41\\
        Highway 1 & 29 & 47 & 0 & 1\\
        Highway 2 & 1 & 21 & 0 & 1\\
        Highway 3 & 63 & 220 & 0 & 0\\
        \hline
        \hline
    \end{tabular}
    \label{table: utilization}
\end{table}

\begin{table}[t]
    \scriptsize
    \centering
    \caption{Number of Days the Charging Sites experienced $P_I > 90\%$.} 
    \label{tab:Idle_90} 
    \includegraphics[width=0.75\linewidth]{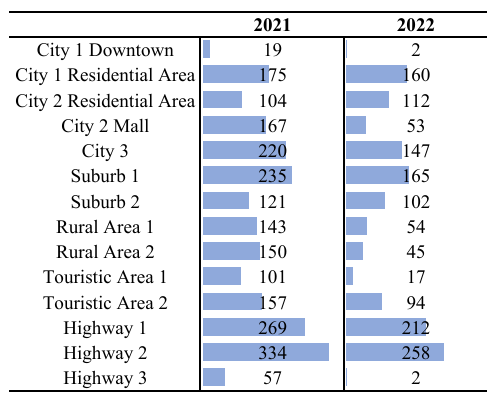}
\end{table}

\begin{figure}[t]
	\centering
    \scriptsize
	\includegraphics[width=0.95\linewidth]{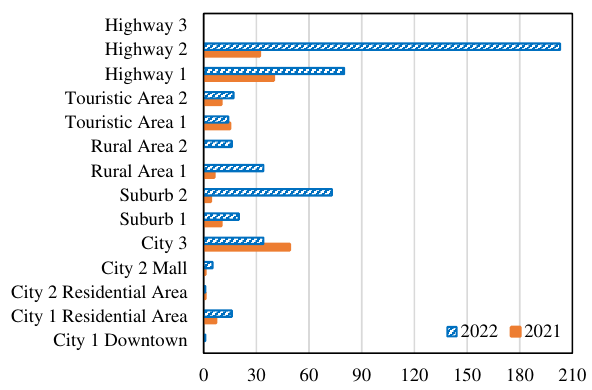}
	\caption{Number of Days with $P_I = 100\%$.}
	\label{fig: Idle_100}
\end{figure}

\begin{figure}[t]
	\centering
    \scriptsize
	\includegraphics[width=0.9\linewidth]{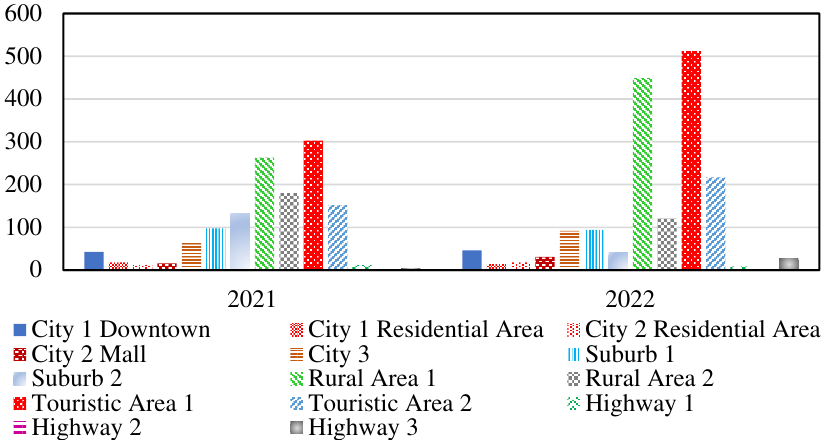}
	\caption{Evolution of $N_{D}$ on the $14$ sites between $2021$ and $2022$.}
	\label{fig: Waited}
\end{figure}

\begin{table}[t]
    \scriptsize
    \centering
    \caption{Blocking probability evolution from $2021$ to $2022$.} 
    \label{tab: block} 
    \includegraphics[width=0.99\linewidth]{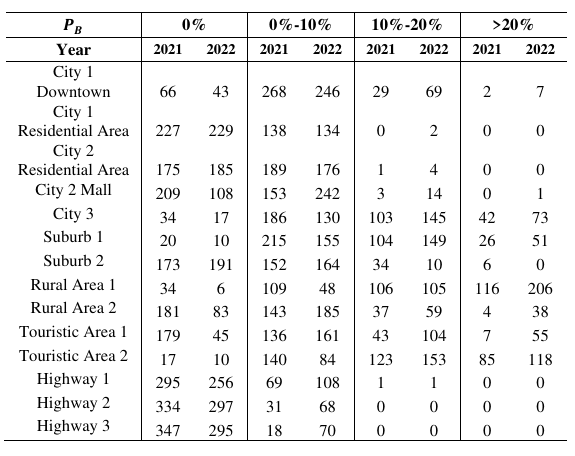}
\end{table}

\subsection{Occupancy, Utilization and Idle Probability:}

The number of EVCSs varies from one to another of the $14$ sites selected from diverse province-wide locations to capture EV charging load variability. For privacy reasons, these sites' actual locations are replaced with generic names as listed in Table \ref{table: sites}. Note that a site with $1$ EVCS, yields $\Omega = U$.

Table \ref{table: occupancy} presents a summary of the number of days where the per-site occupancy belonged to the normal range of $10\%$ to $30\%$ as well as the number of days where that occupancy exceeded $30\%$. Table \ref{table: utilization} on the other hand presents the number of days where the per-site utilization belonged to similar intervals. $\Omega$ and $U$ are presented for both $2021$ and $2022$ to demonstrate the evolution of PCI utilization. Observe that $13$ out of the $14$ sites, witnessed an increase in $\Omega$ and $U$. Particularly, Touristic Area (TA) $2$ and Suburban Area (SA) $1$ exhibited an $\Omega > 30\%$ for more than double the days. Another insight that can be extracted from Table \ref{table: occupancy} and \ref{table: utilization} is that the charging sites next to the Highways $1$ and $2$ are being lightly underutilized with a $U > 30\%$ on zero and one day respectively in $2021$ and $2022$. This could indicate the user's tendency to charge at their origin or destination points rather than spending time charging en route. Also, the charging site on Highway $3$ suffers an $\Omega > 30\%$. However, its $U < 30\%$ throughout all of $2021$ and $2022$. A third observation can be made regarding the charging site in City $2$'s Mall Parking Lot. Even though the number of days where $10\% \leq U \leq 30\%$ doubled, it has never been over-utilized as the number of days where its $U > 30\%$ remains zero. However, this serves as an indicator that another EVCS should be added to that site.

On the other hand, $P_I$ indicates whether or not a charging site is being underutilized. Table \ref{tab:Idle_90} presents the number of days where the $14$ charging sites experienced a $P_{I} > 90\%$. This means that on those days the charging site was vacant and remained idle for $90\%$ of the time; thus, indicating a very low utilization on that day. Table \ref{tab:Idle_90} that charging sites at Highways $1$ and $2$ are severely underutilized. As such, no new EVCSs need to be added at these locations in the near future. However, one important observation is that all sites experienced a drop in $P_{I}$ including Suburb $2$. This means that, even though the overall utilization of these sites decreased from $2021$ to $2022$, it became less likely to find them idle. The shaded bars in Table \ref{tab:Idle_90} represent the percentage of the year during which $P_{I} > 90\%$.

Finally, Figure \ref{fig: Idle_100} presents the number of days during which the charging sites were not used at all (i.e. $P_{I} = 100\%$). This figure can give us three very important observations. The first is that even though certain sites have a high possibility of being idle, the number of days during which no charging occurred on these sites is relatively low. The second observation is on the site at Highway $2$ that had a $P_{I} > 90\%$ for $258$ days of the year but only had $32$ days during which it was not used. This means that although such charging sites will experience low utilization, they are indeed needed almost every day by EV drivers. The third observation was seen at the charging site in City $3$. Although this city experienced higher utilization and occupancy in $2022$, the number of days during which it experienced an $N_{R}$ of zero slightly increased. This type of anomaly represents a trend of more centralized EV charging. It can possibly mean that although user demand on this site increased, this demand was concentrated on certain days. This clearly demonstrates the importance of examining these metrics per day instead of aggregating entire months or entire years together.

\begin{figure}[t]
	\centering
    \scriptsize
	\includegraphics[width=0.95\linewidth]{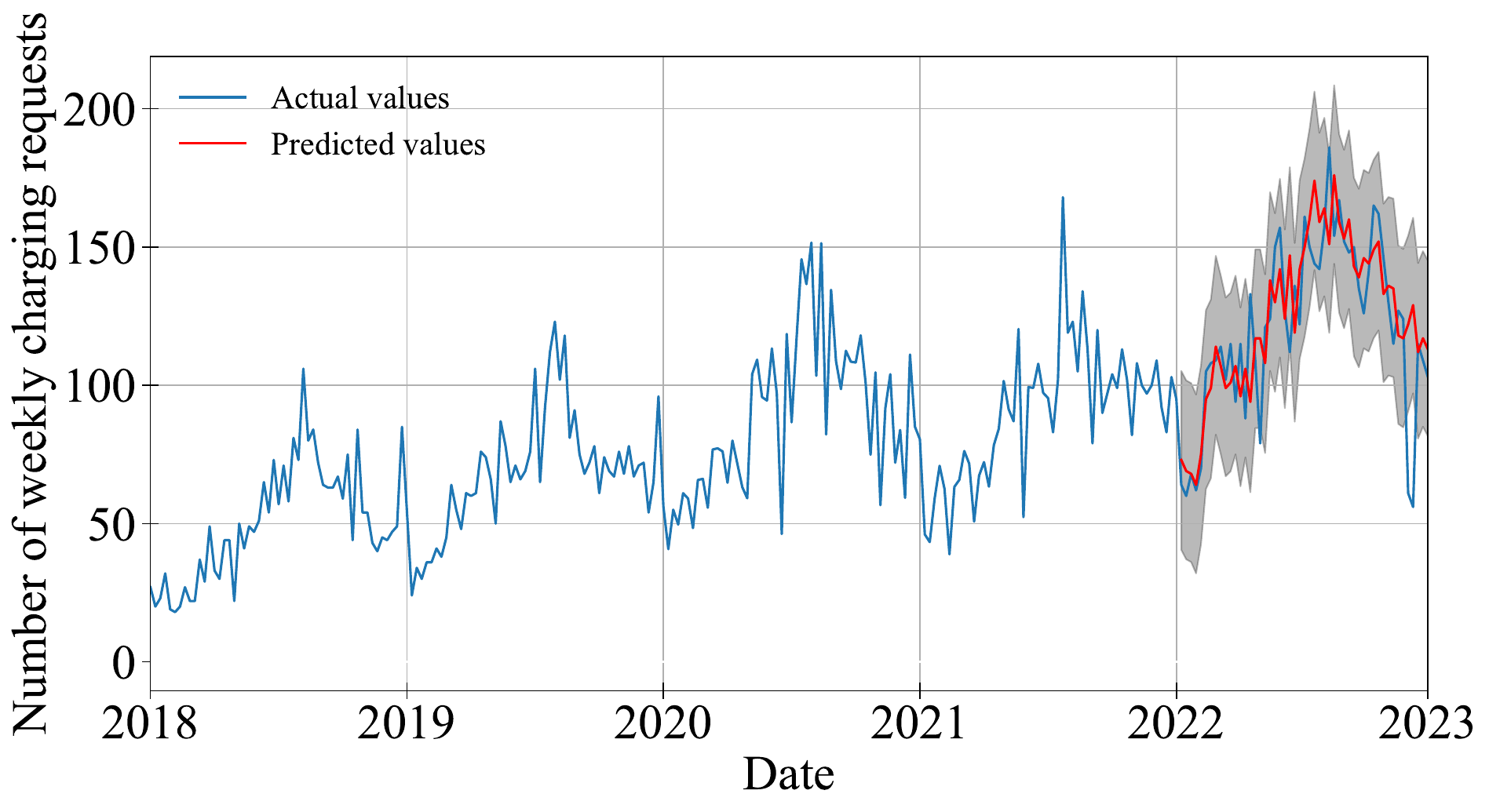}
	\caption{SARIMAX testing accuracy for charging site at Touristic Area 1.}
	\label{fig: SARIMAXtest1}
\end{figure}

\begin{figure}[t]
	\centering
    \scriptsize
	\includegraphics[width=0.95\linewidth]{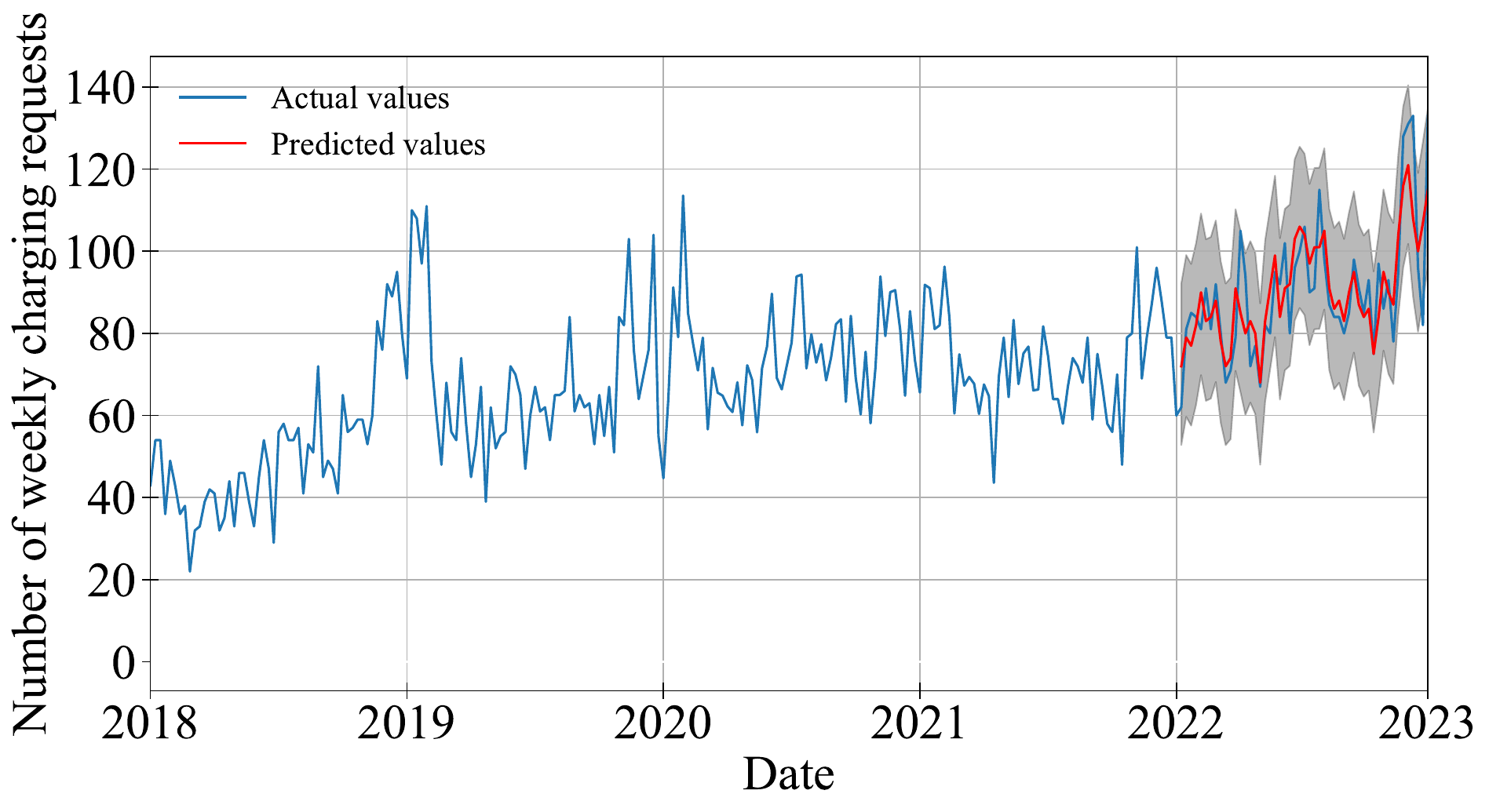}
	\caption{SARIMAX testing accuracy for charging site at City 1 Downtown.}
	\label{fig: SARIMAXtest3}
\end{figure}

\begin{figure}[t]
	\centering
    \scriptsize
	\includegraphics[width=0.95\linewidth]{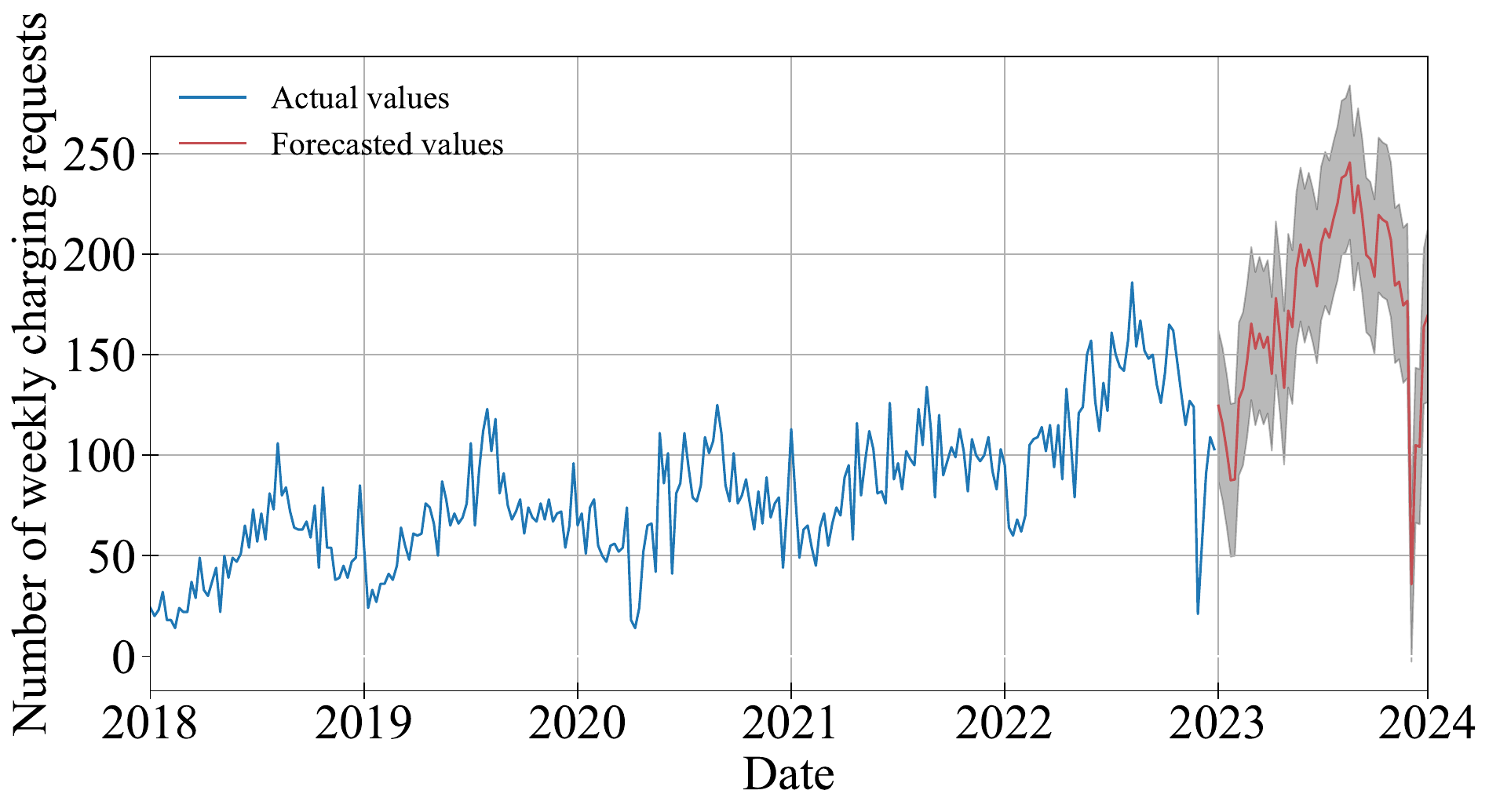}
	\caption{SARIMAX Touristic Area 1's weekly forecasts during $2023$.}
	\label{fig: SARIMAXforecast1}
\end{figure}

\begin{figure}[t]
	\centering
    \scriptsize
	\includegraphics[width=0.95\linewidth]{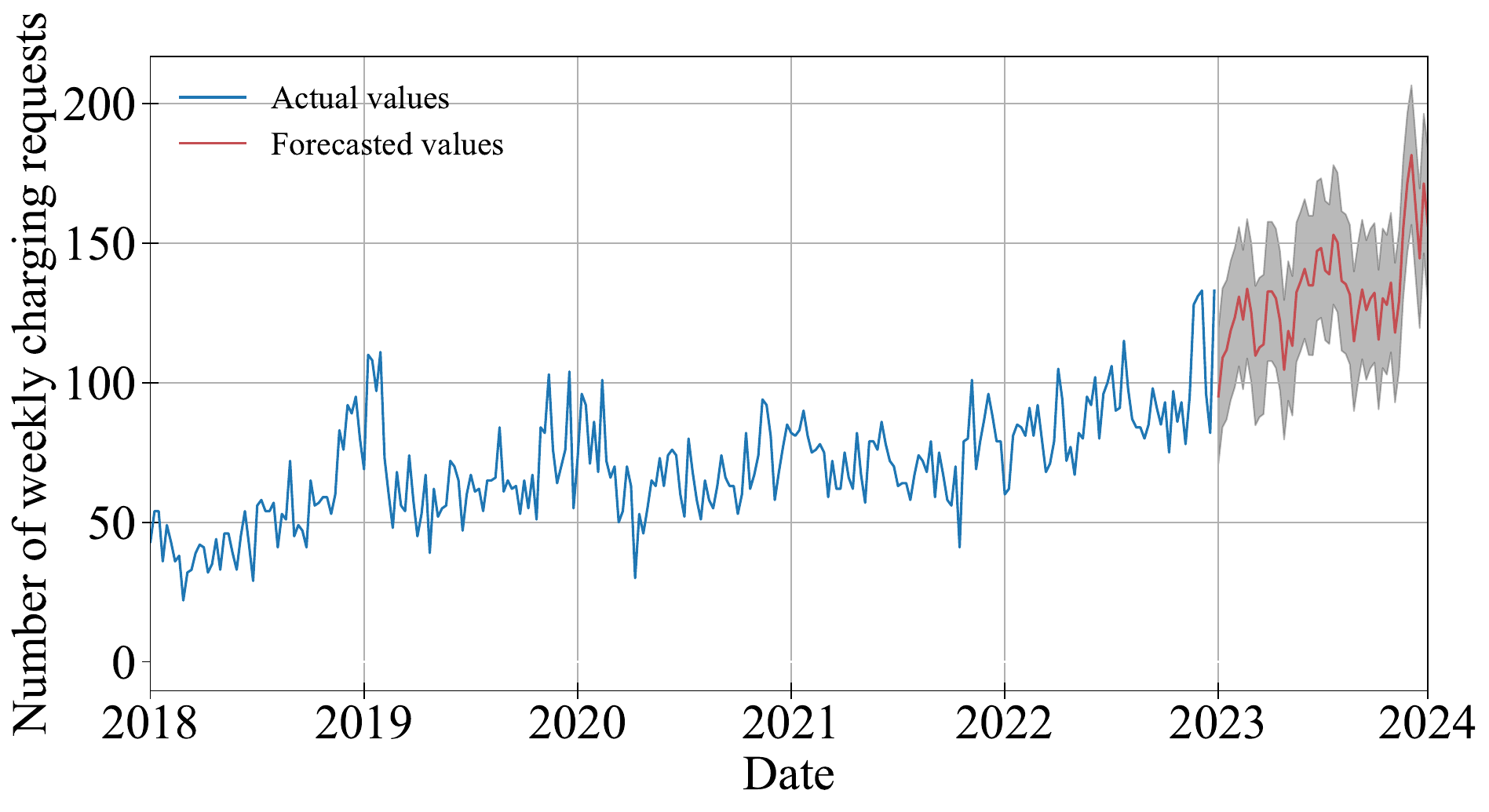}
	\caption{SARIMAX City 1 Downtown's weekly forecasts during $2023$.}
	\label{fig: SARIMAXforecast3}
\end{figure}

\subsection{Blocking Probability:}

Another important metric studied at these $14$ sites is the blocking probability, $P_{B}$. Due to the sensitivity of this metric, it is presented in Table \ref{tab: block} in steps of $10\%$. $P_{B} = 0$ means that a site was not blocked on a given day. A close inspection of Table \ref{tab: block} evides that the charging site next to Highway $1$ was never blocked for $295$ and $256$ days in $2021$ and $2022$ respectively; hence, confirming that EV drivers were less likely to use this specific site. Also, the number of days during which $P_{B} = 0$ for most of the $14$ sites decreased between $2021$ and $2022$. This goes in line with the results reported in Table \ref{table: occupancy} and Table \ref{table: utilization} indicating an increase in these sites' $U$ and $\Omega$ respectively in $2022$. Unexpected, however, is $P_B$ pertaining to Rural Area $1$'s site. While in $2021$ this site had a $P_{B} \geq 20\%$ for $116$ days, this number almost doubled to reach $206$ days in $2022$. That is, an arriving EV at this site had a $20\%$ chance of not finding an available EVCS on $206$ out of $365$ days that year. This is a clear indication that an additional EVCS needs to be added within that site's region. While the charging site in City $1$ Downtown also had a low chance of having $P_{B}= 0$, its performance was still acceptable in $2022$ as most of the days experienced  $0\% \leq P_{B} \leq 10\%$. These results are also reflected by $N_D$ as discussed below.

\subsection{Delayed EVs:}

Recall that $N_{D}$ represents the number of EVs that had to wait at a blocked charging site; hence, the direct relationship between the $P_{B}$ and $N_{D}$. This is revealed in Figure \ref{fig: Waited}, which illustrates the evolution of $N_{D}$ between $2021$ and $2022$ at the $14$ sites. The figure indicates that the site in Touristic Area 1 experienced an increase in $N_D$ by $67\%$ meaning that the service will quickly deteriorate at this site in the coming year if the PCI in that area is not properly expanded. The results pertaining to Highway $1$'s charging site are consistent with their counterparts for $U$, $\Omega$ and $P_B$ reported in their above-indicated respective tables. In confirmation of this site's low $U$ and $P_B$, only $9$ and $12$ EVs had to wait to receive service there for during the entire years $2021$ and $2022$ respectively. Another important observation is the decrease in $N_{D}$ in the Suburban Area $2$ from $134$ EVs in $2021$ to $41$ EVs in $2022$. This result is consistent with the drop in this site's $P_B$.

\begin{figure*}[t]
    \centering
    \subfigure
        {\includegraphics[width=0.32\linewidth]{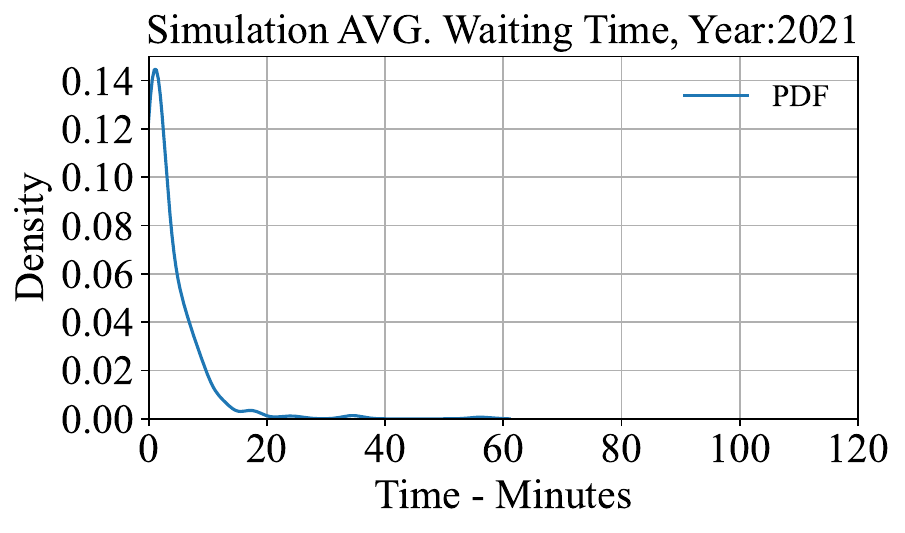}}
    \subfigure
        {\includegraphics[width=0.32\linewidth]{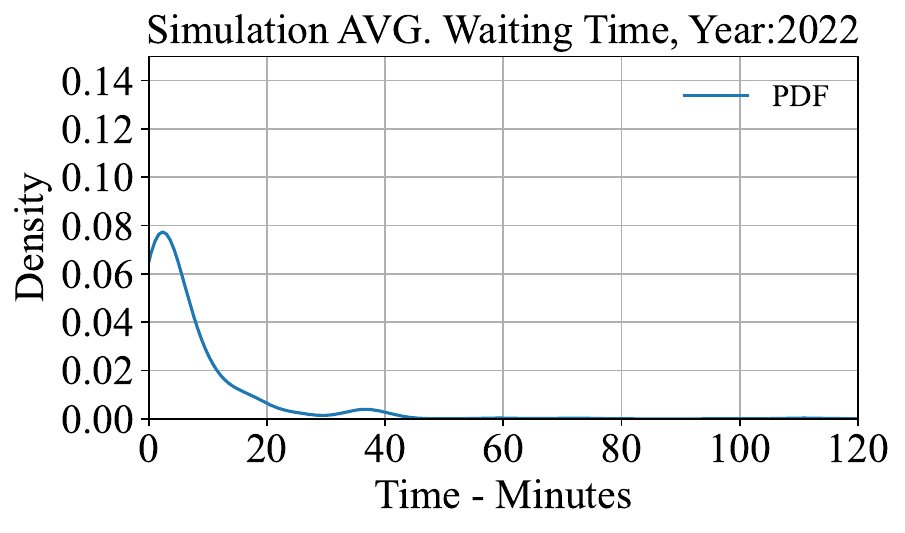}}
    \subfigure
        {\includegraphics[width=0.32\linewidth]{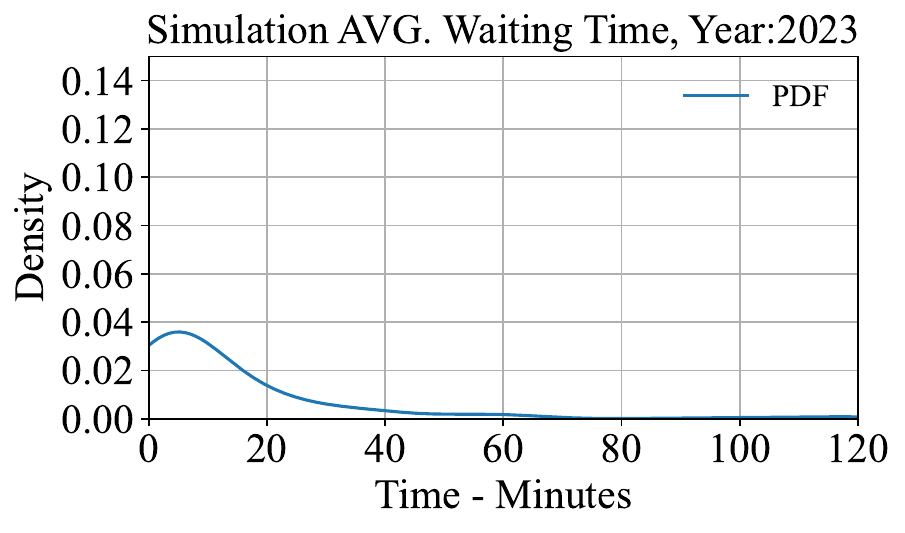}}
    \caption{Evolution of $\overline{W}$'s distribution during $2021$ (left), $2022$ (center), and $2023$ (right) for the site at Touristic Area 1.}
    \label{fig:distrib}
   % \vspace{-0.5cm}
\end{figure*}

\begin{table*}
\centering

\caption{Tabulated Quarterly Forecast Results.}
\begin{tabular}{||c||c|c|c|c||}
    \hline
    \bf{Quarter} & \bf{Touristic Area 1} & \bf{City 1 DownTown} & \bf{Rural Area 1} & \bf{City 2 Mall} \\ 
    \hline
    \hline
    2023 Q1 & $1539$ & $1452$ & $526$ & $1098$ \\
    2023 Q2 & $2263$ & $1709$ & $811$ & $895$ \\
    2023 Q3 & $2841$ & $1772$ & $1278$ & $1008$ \\
    2023 Q4 & $2339$ & $1898$ & $801$ & $1178$ \\
    2024 Q1 & $2382$ & $2260$ & $795$ & $1655$ \\
    2024 Q2 & $3267$ & $2444$ & $1206$ & $1202$ \\
    2024 Q3 & $3927$ & $2462$ & $1740$ & $1422$ \\
    2024 Q4 & $2979$ & $2500$ & $1010$ & $1526$ \\
    \hline
    \hline   
    \end{tabular}
    \label{tab: quartely_results}
\end{table*}

\begin{table}
\centering
\caption{Tabulated Model Parameters.}
\begin{tabular}{||c||c|c|c|c|c|c|c||}
    \hline
    \bf{Site Location} & \bf{p} & \bf{d} & \bf{q} & \bf{P} & \bf{D} & \bf{Q} & \bf{Model MAPE}  \\ 
    \hline
    \hline
    Touristic Area 1 & 0 & 0.6 & 1 & 0 & 1 & 1 & $12.12\%$ \\
    City 1 Downtown 1 & 1 & 0.8 & 1 & 0 & 1 & 1 & $8.15\%$ \\
    City 2 Mall 1 & 1 & 0.8 & 0 & 1 & 1 & 0 & $14.28\%$ \\
    Rural Area 1 & 0 & 0.4 & 1 & 0 & 1 & 1 & $10.83\%$ \\
    
    \hline
    \hline   
    \end{tabular}
    \label{tab: model_pqd}
\end{table}

\subsection{SARIMAX Model Testing and Forecast:}

The SARIMAX model presented earlier in Section V-C is now used to predict future values of $N_{R}$ for $4$ of the $14$ sites examined in this manuscript, namely, Touristic Area 1, City 1 Downtown, Rural Area 1, and City 2 Mall. The model is trained using the $2018$ through $2021$ data pertaining to each of these charging sites individually and then tested against the $2022$ data. The grid search algorithm is then utilized to evaluate the best parameters that would achieve the most accurate results when validated against the $2022$ data. The \texttt{p}, \texttt{q}, and \texttt{d} parameters of these models are presented in Table \ref{tab: model_pqd} for the reader's convenience. Figures \ref{fig: SARIMAXtest1} and \ref{fig: SARIMAXtest3} constitute tangible proofs of the model's forecast accuracy for the sites at Touristic Area 1 and City 1 Downtown. The SARIMAX model was able to accurately forecast the charging demand while very closely following the behavior of the actual time series. The shaded region in the figure also indicates that future demands are predicted with $99\%$ confidence. The fine-tuned models are now used to forecast the demand for the entire year of $2023$ for these $4$ sites. textcolor{blue}{Additionally, these models were also used to generate a forecast for the weekly number of requests for a second year, $2024$.} Due to space limitation, only the forecast results of textcolor{blue}{the year 2023 of} $2$ sites are presented in Figure \ref{fig: SARIMAXforecast1} and Figure \ref{fig: SARIMAXforecast3}. However, the total quarterly number of requests for the $4$ sites for the entire $2$-year period is presented in Table \ref{tab: quartely_results}. Additionally, the accuracy evaluation metrics pertaining to the adopted model herein are presented in Table \ref{tab:my_label}. For the sake of completeness, the accuracy of this model is compared to those of 5 other counterparts proven in the literature to outperform the presently suggested SARIMAX model with fractional parameters. Precisely, the results of the SARIMAX model presented above are compared with two ETS models, with and without seasonality, an LSTM model, an ARIMA model without seasonality, and finally with an SARIMA model without an exogenous variable. Table \ref{tab:my_label} clearly demonstrates the superiority of the presented SARIMAX model with fractional parameters in terms of forecast accuracy. Here, it is very interesting to observe the relatively poor performance of the LSTM model despite notable efforts to optimize its performance as well as the inclusion of the exogenous parameter as a feature for the LSTM's NN. This is a consequence of the relatively young age of the charging infrastructure, which would severely limit the success of training a Deep Learning algorithm due to the limited number of training samples in a $4$-year period ($201$ weekly values per site). Also, traditionally, such forecasts rely on extensive historical information spanning tens of years (such as stock market forecasts).

\begin{table*}
\centering

\caption{Performance metrics for different models.}
\begin{tabular}{||c||c||c|c|c|c||}
    \hline
    \bf{Site} & \bf{Model} & \bf{MSE} & \bf{RMSE} & \bf{MAPE} & \bf{MAE} \\
    \hline
    \hline
    & ETS Without Seasonality & $754.32$ & $27.46$ & $21.04\%$ & $21.53$ \\
    & ETS With Seasonality & $530.74$ & $23.04$ & $17.65\%$ & $18.67$ \\
    & LSTM & $690.50$ & $26.28$ & $19.11\%$ & $19.10$ \\
    \bf{Touristic Area 1} & ARIMA & $486.29$ & $22.05$ & $17.28\%$ & $17.71$ \\
    & SARIMA & $621.69$ & $24.93$ & $18.78\%$ & $19.40$ \\
    & SARIMAX & $423.76$ & $20.59$ & $12.12\%$ & $15.06$ \\
    \hline
    \hline
    & ETS Without Seasonality & $688.89$ & $26.24$ & $22.82\%$ & $21.77$ \\
    & ETS With Seasonality & $276.84$ & $16.79$ & $12.55\%$ & $11.78$ \\
    & LSTM & $375.68$ & $19.38$ & $17.56\%$ & $15.39$ \\
    \bf{City 1 Downtown} & ARIMA & $187.39$ & $13.69$ & $11.55\%$ & $10.84$ \\
    & SARIMA & $189.44$ & $13.76$ & $11.33\%$ & $10.60$ \\
    & SARIMAX & $90.04$ & $9.49$ & $8.15\%$ & $7.57$ \\
    \hline
    \hline
    \end{tabular}
    \label{tab:my_label}
\end{table*}

\subsection{Waiting Time Evolution:}

After forecasting the number of charging requests for $2023$ the \textit{M}/\textit{G}/$k$ simulator presented in Section V-A is used to study the evolution of the waiting time on the $4$ selected sites throughout the years $2021$ (including the filled Gap), $2022$, and $2023$ (forecasts). The daily number of requests is first extracted from the weekly forecast values. Then, these values are used to determine the EV arrival rates to the charging site. Subsequently, extensive simulations are performed with $25,000$ EV arrivals per simulation round generated using the determined arrival rates for all of the $365$ days corresponding to each year. This guarantees the achievement of at least $95\%$ confidence interval and reduces the impact of any outliers.

The results of the three years’ simulations for Touristic Area 1 are presented in Figure \ref{fig:distrib}. This figure demonstrates how the probability of low waiting times is decreasing while the probability of experiencing high waiting times is increasing year after year. This is demonstrated by the bulkier tale of the distribution and the more frequent $\overline{W}$ values above $60$ minutes. To put things into perspective, the probability of experiencing a waiting time of zero is $20.4\%$ and $9.3\%$ in $2021$ and $2022$ respectively. This value drops to $3.8\%$ in $2023$. Additionally, the average waiting time in $2021$ was $3.49$ minutes and rises to $6.73$ minutes and $14.68$ minutes in $2022$ and $2023$ respectively. Additionally, $99\%$ of the anticipated waiting times are within $30$ minutes, $40$ minutes, and $110$ minutes in $2021$, $2022$, and $2023$ respectively. The statistics related to the $\overline{W}$ results of the $4$ analyzed sites are presented in Table \ref{table: Wait_stats}.

\begin{table}[t]
    \centering
    \caption{\textsc{Waiting Time Statistics for 2021, 2022, and 2023}}
    \begin{tabular}{c | c |c | c}\hline
        \multicolumn{1}{c}{2021} & \multicolumn{1}{c}{Average} & \multicolumn{1}{c}{Probability $\overline{W}=0$} & \multicolumn{1}{c}{99\% Interval}\\\hline
        City 1 Downtown & 2.22 min & 26\% & 6 min\\
        City 2 Mall & 1 min & 50\% & 2 min\\
        Rural Area 1 & 19.95 min & 0\% & 36 min\\
        Touristic Area 1 & 3.49 min & 20.4\% & 30 min\\\hline
        \multicolumn{1}{c}{2022} & \multicolumn{1}{c}{Average} & \multicolumn{1}{c}{Probability $\overline{W}=0$} & \multicolumn{1}{c}{99\% Interval}\\\hline
        City 1 Downtown & 3.68 min & 10.5\% & 10 min\\
        City 2 Mall & 1.3 min & 44\% & 3 min \\
        Rural Area 1 & 24.98 min & 0\% & 43 min\\
        Touristic Area 1 & 6.73 min & 9.3\% & 40 min\\\hline
        \multicolumn{1}{c}{2023} & \multicolumn{1}{c}{Average} & \multicolumn{1}{c}{Probability $\overline{W}=0$} & \multicolumn{1}{c}{99\% Interval}\\\hline
        City 1 Downtown & 11.32 min & 4.9\% & 24 min\\
        City 2 Mall & 2.9 min & 22.6\% & 7 min\\
        Rural Area 1 & 35.36 min & 0\% & 85 min \\
        Touristic Area 1 & 14.68 min & 3.8\% & 110 min\\\hline
    \end{tabular}
    \label{table: Wait_stats}
\end{table}

\subsection{Case Study: Charging Site at Touristic Area $2$:}

Finally, a case study related to Touristic Area $2$ is considered. By the end of $2021$, this site contained a total of $9$ Level $2$ EVCSs and $2$ Level 3 EVCs. The site under study contains $1$ of these Level $3$ EVCSs. This region experienced a rapid deployment of Level $2$ EVCS rising to reach a total of $33$ Level $2$ EVCSs by the end of $2022$. However, no new Level $3$ EVCSs were added in $2023$ to this region. Based on the earlier conducted analysis, the utilization of the specific site of interest kept increasing in $2023$ despite the addition of $24$ new Level $2$ EVCSs within a very short distance. This site's $P_B$ and $\overline{W}$ also increased. This confirms the fact that, even though the Level $2$ PCI was greatly expanded in this region, it failed to catch up with the increasing EV charging demands; hence, the extreme importance of utilizing the herein presented methodology to determine the sites worthy of new Level $3$ EVCS deployments for the purpose of maintaining adequate QoE performance and supporting the growing number of EVs and their charging demands.

\section{Conclusion} \label{sec: Conclusion}

This is the first data-driven study that develops a comprehensive set of metrics to evaluate the EV PCI performance of future smart cities. Also, the herein-developed forecast model accurately predicts the evolution of charging requests at a given site. The performance pertaining to $14$ representative sites was analyzed through a close examination of the evolution of the occupancy, utilization, blocking probability, and the number of waiting EVs. A custom-built simulator is then used to estimate the average waiting time experienced by EVs during $2021$ and $2022$ as well as the predicted waiting time given the $2023$ predicted charging demand. This study demonstrates the necessity of expanding the EV PCI in order to satisfy the ever-increasing charging demand and maintain acceptable consumer-perceived QoE levels.\\


\begin{thebibliography}{1}

%\bibitem{France2030}
%    Elysée: "France 2030: La France de demain commence aujourd'hui," URL: https://www.elysee.fr/emmanuel-macron/france2030.

%\bibitem{MITNews}
%    N. W. Stauffer, "China's Transition to Electric Vehicles," MIT News, 2021, URL: https://news.mit.edu/2021/chinas-transition-electric-vehicles-0429.

%\bibitem{PROTOCOL}
%    Z. Yang, "China Plans Charging Infrastructure For 20 Million EVs," Protocol, 2022, URL: https://www.protocol.com/bulletins/china-infrastructure-plan-ev.

%\bibitem{HARVARD}
%    A. Sillman, "A New Automotive Era? Promoting Electric Vehicles in The Biden Administration," Harvard Law School $-$ Environmental \& Energy Law Program, 2021, URL: https://eelp.law.harvard.edu/2021/03/new-auto-era-biden/

%\bibitem{REUTERS}
%    D. Shepardson, (et. al.), "Biden Seeks to Make Half of New U.S. Auto Fleet Electric By 2030," REUTERS, 2021, URL: https://www.reuters.com/business/autos-transportation/biden-set-target-50-evs-by-2030-industry-backs-goal-2021-08-05/

%\bibitem{CADTRANS}
%    Transport Canada, "Minister of Transport Announces The Expansion of The Incentives for Zero-Emission Vehicles Program," Transport Canada News Release, 2022, URL: https://www.canada.ca/en/transport-canada/news/2022/04/minister-of-transport-announces-the-expansion-of-the-incentives-for-zero-emission-vehicles-program.html

\bibitem{IEA}
    IEA, "Global EV Outlook 2022", IEA, Paris, 2022, URL: https://www.iea.org/reports/global-ev-outlook-2022/

\bibitem{AFID}
    I. Ertug, "Revision of the Directive on Deployment of Alternative Fuels Infrastructure," \emph{A European Green Deal}, 2023, URL:https://www.europarl.europa.eu/legislative-train/theme-a-european-green-deal/file-revision-of-the-directive-on-deployment-of-alternative-fuels-infrastructure

\bibitem{BernardEtAl}
    M. R. Bernard, (\textit{et. al}), "Assessing Charging Infrastructure Needs in Quebec," \emph{ICCT}, Working Paper 2021-45, 2022.

\bibitem{UnterluggauerEtAl}
    T. Unterluggauer, (\textit{et. al.}, "Electric Vehicle Charging Infrastructure Planning For Integrated Transportation and Power Distribution Networks: A Review," \emph{eTransportation}, 12, 2022.

\bibitem{VashisthEtAl}
    S. Vashisth, (\textit{et. al.}), "Multi-stage Planning of Fast Charging Stations For PEVs Using Traffic-based
    Approach," \emph{Sust. Ener., Grids and Net.}, 30, 2022.

\bibitem{HafezEtAl}
    O. Hafez, (\textit{et. al.}), "Queueing Analysis Based PEV Load Modeling Considering Battery Charging Behavior and Their Impact on Distribution System Operation," \emph{IEEE TSG}, 9:1, 2016.

\bibitem{WRTTS}
    Waterloo Region Transportation Tomorrow Survey, 2016, URL: http://dmg.utoronto.ca/transportation-tomorrow-survey/tts-reports

\bibitem{YiEtAl}
    Z. Yi, (\textit{et. al.}), "Electric Vehicle Demand Estimation and Charging Station Allocation Using Urban Informatics," \emph{Transp. Res. Part D: Trans. and Env.}, 106, 2022.

\bibitem{PageEtAl}
    L. Page, (\textit{et. al.}), "The PageRank Citation Ranking: Bringing Order to The Web," \emph{Stanford InfoLab}, TR, 1999, URL: \url{http://ilpubs.stanford.edu:8090/422/}

\bibitem{OrzechowskiEtAl}
    A. Orzechowski, (\textit{et. al.}), "A Data-driven Framework For Medium-Term Electric Vehicle Charging Demand Forecasting," ELSEVIER Energy and AI, 14, 2023.
    
\bibitem{KabirEtAl}
    M. E. Kabir, (\textit{et. al.}), "Demand-Aware Provisioning of Electric Vehicles Fast Charging Infrastructure," \emph{IEEE TVT}, 69:7, 2020.

\bibitem{KhodayarEtAl}
    M. E. Khodayar, (\textit{et. al.}), "Hourly Coordination of Electric Vehicle Operation and Volatile Wind Power Generation in SCUC," \emph{IEEE TSG}, 3:3, 2012.

\bibitem{CaoEtAl}
    Y. Cao, (\textit{et. al.}), "An Optimized EV Charging Model Considering TOU Price and SOC Curve," \emph{IEEE TSG}, 3:1, 2012.

\bibitem{AntounEtAl}
    J. Antoun, (\textit{et. al.}), "A Data Driven Performance Analysis Approach For Enhancing The QoS of Public Charging Stations," \emph{IEEE T-ITS}, 23:8, 2022.
    
\bibitem{AriasEtAl}
    M. B. Arias, (\textit{et. al.}), "Electric Vehicle Charging Demand Forecasting Model Based on Big Data Technologies," \emph{ELSEVIER AP-ENERGY}, 183, 2016.

\bibitem{MuEtAl}
    Y. Mu, (\textit{et. al.}), "A Spatial-Temporal Model For Grid Impact Analysis of Plug-in Electric Vehicles,", \emph{ELSEVIER AP-ENERGY}, 114, 2014.

\bibitem{WangEtAl}
    H. Wang, (\textit{et. al.}), "Load Characteristics Of Electric Vehicles In Charging And Discharging States And Impacts On Distribution Systems," \emph{Proc. SUPERGEN}, China, 2012.

\bibitem{QianEtAl}
    K. Qian, (\textit{et. al.}), "Modeling Of Load Demand Due To EV Battery Charging In Distribution Systems," \emph{IEEE TPS}, 26:2, 2011.

\bibitem{LojowskaEtAl}
    A. Logowska, (\textit{et. al.}), "From The Transportation Patters To Power Demand: Stochastic Modeling Of Uncontrolled Domestic Charging Of Electric Vehicles," \emph{Proc. IEEE PES}, United States, 2011.

\bibitem{MaEtAl}
    T.-Y. Ma (\textit{et. al.}), "Multistep Electric Vehicle Charging Station Occupancy Prediction Using Hybrid LSTM Neural Networks," \emph{Energy}, 244, 2022.

\bibitem{BergstraEtAl}
    J. Bergstra, (\textit{et. al.}), "Alogrithms For Hyper-parameter Optimization," \emph{Proc. ACM NIPS}, United States, 2011.

\bibitem{LinoffEtAl}
    G. S. Linoff, (\textit{et. al.}), "Data Mining Techniques For Marketing, Sales and Customer Support," \emph{Wiley}, 2011.

\bibitem{SmolaEtAl}
    A. J. Smola, (\textit{et. al.}), "A Tutorial On Support Vector Regression," \emph{SPRINGER Stat. \& Comp.}, 14, 2004.

\bibitem{Ho}
    T.-K. Ho, "Random Decision Forest," \emph{Proc. IEEE ICDAR}, Canada, 1995.

\bibitem{FreundEtAl}
    Y. Freund, (\textit{et. al.}), "A Decision-Theoretic Generalization Of On-Line Learning And An Application To Boosting," \emph{J. Comp. Sys. Sci.}, 55:1, 1997.

\bibitem{AminiEtAl}
    M. H. Amini, (\textit{et. al.}), "ARIMA-based Decoupled Time Series Forecasting of Electric Vehicle Charging Demand For Stochastic Power System Operation," \emph{ELSEVIER EPSR}, 140, 2016.

\bibitem{ManujithEtAl}
    S. K. Manujith, (\textit{et. al.}), "Comparative Analysis of Deep Learning Models For Electric Vehicle Charging Load Forecasting," \emph{J. Inst. of Eng. India: Series B}, 104, 2023.






























\bibitem{AminiEtAl}
    M. H. Amini, (\textit{et. al.}), "ARIMA-based Decoupled Time Series Forecasting of Electric Vehicle Charging Demand For Stochastic Power System Operation," \emph{ELSEVIER EPSR}, 140, 2016.

\bibitem{ManujithEtAl}
    S. K. Manujith, (\textit{et. al.}), "Comparative Analysis of Deep Learning Models For Electric Vehicle Charging Load Forecasting," \emph{J. Inst. of Eng. India: Series B}, 104, 2023.

\bibitem{Chatfield}
Ken Chatfield, et. al. "The devil is in the details: an evaluation of recent feature encoding methods." BMVC. Vol. 2. No. 4. 2011.

\bibitem{encod}
B. Roy, "All about Categorical Variable Encoding," Towards Data Science, https://towardsdatascience.com/all-about-categorical-variable-encoding-305f3361fd02


\bibitem{sarimax}
McLeod, A. I., \& Hipel, K. W. (1981). Seasonal time series models in engineering and environmental studies. Journal of the American Statistical Association, 76(374), 31-36.
\bibitem{arima}
B. Artley, "Time Series Forecasting with ARIMA , SARIMA and SARIMAX," Towards Data Science, 2022.
\bibitem{AR}
J. Fattah, "Forecasting of demand using ARIMA model," International Journal of Engineering Business Management, vol. 10, 2018. 

\end{thebibliography}
\end{document}